\documentclass{article}

\usepackage[inkscapearea=page]{svg}
\usepackage[utf8]{inputenc} 
\usepackage{threeparttable}
\usepackage[T1]{fontenc}    
\usepackage{subcaption}
\usepackage{microtype}      
\usepackage{mathtools}
\usepackage{adjustbox}
\usepackage{booktabs}       
\usepackage{amsfonts}       
\usepackage{nicefrac}       
\usepackage{graphicx}
\usepackage{multirow}
\usepackage{rotating}
\usepackage{amsfonts}
\usepackage{amsmath}
\usepackage{wrapfig}
\usepackage{amssymb}
\usepackage{amsthm}
\usepackage[dvipsnames,svgnames]{xcolor}         
\usepackage{float}
\usepackage{url}            
\usepackage{bbm}
\usepackage{bm}
\usepackage{enumitem}

\usepackage{hyperref}

\usepackage[symbol]{footmisc}

\usepackage[nonatbib, main, final]{neurips_2026}
\usepackage[numbers]{natbib}

\usepackage[capitalize,noabbrev]{cleveref}

\theoremstyle{plain}
\newtheorem{theorem}{Theorem}[section]
\newtheorem{proposition}[theorem]{Proposition}

\theoremstyle{definition}
\newtheorem{definition}[theorem]{Definition}

\newtheorem{remark}[theorem]{Remark}

\usepackage[textsize=tiny]{todonotes}

\newcommand{\overlinehirescam}[1]{\overline{\mathcal{A}}_{#1}}
\newcommand{\hirescam}[1]{\mathcal{A}_{#1}}
\newcommand{\contcam}[0]{\mathcal{C}}
\newcommand{\reconcam}[1]{\contcam^{ag}_{#1}}
\newcommand{\contcamidx}[3][]{\contcam_{(#2, #3)#1}}

\newcommand{\rvd}[0]{\mathbf{d}}
\newcommand{\rvv}[0]{\mathbf{v}}
\newcommand{\rvx}[0]{\mathbf{x}}
\newcommand{\rvy}[0]{\mathbf{y}}

\newcommand{\rmA}[0]{\mathbf{A}}
\newcommand{\rmT}[0]{\mathbf{T}}
\newcommand{\rmX}[0]{\mathbf{X}}

\title{Improving Context-Shift Robustness of Convolutional Networks via Context-Regularized Cross-Entropy}

\author{%
  Jinen Setpal \\
  Elmore Family School of Electrical and Computer Engineering \\
  Purdue University \\
  \texttt{jsetpal@purdue.edu} \\
  \And
  Chaoyue Liu \\
  Elmore Family School of Electrical and Computer Engineering \\
  Purdue University \\
  \texttt{cyliu@purdue.edu} \\
}

\begin{document}

\maketitle

\begin{abstract}
    Image classification models often encounter objects in diverse and unpredictable contexts at test time, which may differ substantially from those seen during training. This phenomenon, termed {\it context shift}, exposes a key fragility in neural networks: when models rely on non-causal contextual cues for prediction, shifts in context can lead to significant performance degradation.
    In this paper, we propose a novel training framework to improve the robustness of convolutional neural networks under context shift. Our approach leverages spatial feature attribution to guide models toward making predictions that rely less on contextual regions and more on object-relevant features. As a first step, recognizing a theoretical limitation of existing feature attribution methods we introduce a new variant, {\it ContrastiveCAMs}, which produces more faithful attribution maps of model predictions.
    Building on ContrastiveCAMs, we further propose {\it Context-Regularized Cross-Entropy} (CR-CE), a modification of the standard cross-entropy loss that regularizes the model’s attention by suppressing the influence of contextual regions, thereby improving robustness to context shift.
    We evaluate the effectiveness of our approach on several medium-to-large scale datasets (Waterbirds, Spawrious, ImageNet/ImageNet-BG, Hard-ImageNet), and report consistent improvements in context-shift robustness. 
\end{abstract}

\section{Introduction}
\label{sec:introduction}

The vast applications of convolutional neural networks in safety-critical domains such as medical imaging \citep{sarvamangala2022convolutional, rajpurkar2017chexnet}, forensic investigation \citep{murthy2023ai} and self-driving \citep{kim2017interpretable} make trustworthy and robust predictions essential for large-scale adoption.
It has been demonstrated that existing neural networks, despite achieving remarkable performance on large-scale datasets, 
heavily rely on supporting context (i.e., background area within the images) to determine their predictions \citep{singla2022salient, lynch2023spawrious, Piekarek_2025_ICCV}.
This sensitivity to surrounding context can degrade performance when the same object appears in a different environment \citep{wang2023feature, geirhos2020shortcut, bassi2024improving, boland2024there}, whether due to novel object-context combinations in the test data or adversaries manipulating target-agnostic regions to mislead the model.

\begin{figure}
    \centering
    \includegraphics[width=\textwidth]{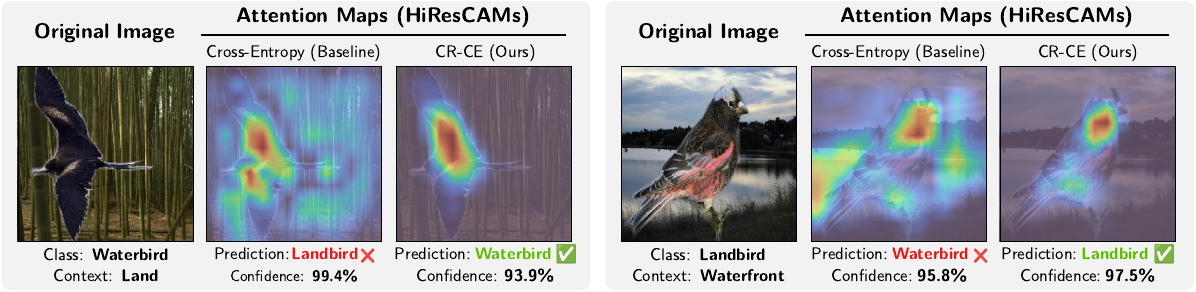}
    \caption{{\bf Context Contributions Can Mislead Predictions.}
    Spatial feature attribution maps reveal that cross-entropy trained models rely heavily on context, causing misclassifications in novel environments.
    Our approach encourages focus on the object thereby improving context-shift robustness.}
    \label{fig:overview}
\end{figure}

We highlight this concern in Figure \ref{fig:overview} using the Waterbirds-100 dataset \citep{sagawa2019distributionally} which only has the object-context combinations of {\it waterbird-waterfront} and {\it landbird-land}.
When a {\it waterbird} is shown in the context of {\it land}, the model predicts the class {\it landbird} with high confidence, due to the influence of the context dominating over the influence of the object.
The converse also occurs when a {\it landbird} appears in a {\it waterfront} context.
This concerning lack of robustness has been observed across safety-critical applications spanning the diagnosis of skin cancer \citep{nauta2021uncovering}, pneumothorax \citep{saab2022reducing}, as well as in the generalization capabilities of robot foundation models \citep{xing2025shortcut}.

In this paper, we aim to improve the context robustness of convolutional neural networks by leveraging spatial feature attributions. We begin by noting a theoretical limitation of existing feature attribution methods: they permit an arbitrary constant shift applied uniformly across all output logits, which undermines their ability to provide faithful explanations. As a result, the attributed contributions from context and object regions may not accurately reflect their true influence, as they can be distorted by this constant shift. We address this limitation by proposing a new variant  {\it ContrastiveCAMs}, based on a prior method, HiResCAMs \citep{draelos2020use}. We further theoretically confirm that {\it ContrastiveCAMs} provide faithful explanations to the neural networks final predictions.

To address sensitivity to context shifts, we propose a new loss function, {\it Context-Regularized Cross-Entropy} (CR-CE), which leverages the proposed feature attribution method, {\it ContrastiveCAMs}, to encourage the model to focus on the primary object by suppressing contributions from the surrounding context. This approach requires object-context segmentation during training (crucially, with no segmentation required during inference), which can be readily obtained using existing pretrained segmentation models such as SAM3 \cite{carion2026sam} and Masked AutoEncoders (MAE) \cite{gao2021luss, he2022masked}.
Notably, the segmentation need not be perfect.
In our experiments on the ImageNet dataset, we use MAEs to generate object-context segmentation.
Experimental results demonstrate that our method effectively reduces the context contribution to model predictions and improves convolutional networks' robustness to context shifts consistently across a broad class of datasets.
Our key contributions are summarized as follows:
\begin{itemize}[leftmargin=*]
    \item We identify a theoretical limitation of existing spatial feature attribution methods that hinders explanation faithfulness.
    We address this limitation by proposing a new method {\it ContrastiveCAMs}.
    \item We propose a new loss function, {\it Context-Regularized Cross Entropy} (CR-CE), that encourages model predictions to focus on the object region rather than the context.
    \item Experimental evaluations demonstrate that the proposed training approach effectively attracts the focus of feature attribution towards the object region, thus enhancing context-shift robustness.
\end{itemize}

\paragraph{Notation.} We denote vectors using bold lowercase letters (e.g., $\rvv$), matrices using uppercase letters (e.g., $M$), and tensors using bold uppercase letters (e.g., $\rmT$), with partial indexing implying selection of the subtensor across the remaining subsequent dimensions (e.g., $\rmT_i \in \mathbb{R}^{b \times c}$ for $\rmT \in \mathbb{R}^{a \times b \times c}$). We use the operator $\odot$ to represent elementwise multiplication, and define $[K] := \{1, 2, \ldots, K\}$.

\subsection{Related Work}

\paragraph{Feature Attribution Methods in Convolutional Networks.}
Class Activation Mappings (CAMs) \citep{zhou2016learning} is a prominent family of methods that identify attributions of each (spatial) feature, including  
GradCAM \citep{selvaraju2017grad}, GradCAM++ \citep{Chattopadhay_2018}, Score-CAM \citep{wang2020score}, and HiResCAM \citep{draelos2020use}. These methods typically compute feature attributions with respect to the model’s output logits, prior to the final softmax activation. However, due to the contrastive nature of the softmax function \citep{oh2025beyond,piland2026diffgradcam}, this can lead to inaccurate attributions for the post-softmax predictions. DiffGradCAM \cite{piland2026diffgradcam} shares a similar approach to ContrastiveCAMs. But it is based on GradCAM and its connection to the model's prediction is not theoretically clear.

\paragraph{Context-Shift Robustness.} One approach to improving robustness is through explanation-guided methods, which leverage feature attribution or saliency maps \citep{Simonyan14a} to direct the model’s attention toward relevant objects and away from contextual distractions in the images \citep{gao2024going}.
This is typically achieved by augmenting the standard cross-entropy loss with additional regularization terms that penalize discrepancies between the model’s feature attribution map and a target explanation \cite{ross2017right,petryk2022gals,rao2023studying,fukui2019attention}.
The regularization term of Right for the Right Reasons (RRR) \citep{ross2017right}  penalizes the gradients of log-prediction w.r.t.~the irrelevant input features. 
GALS \citep{petryk2022gals} regularizes the Grad-CAM of the model’s predictions toward a target feature attribution map generated using training labels and auxiliary vision-language models.
The regularization term of EBL \citep{rao2023studying} penalizes the positive attributions in contextual regions of the feature attribution map.
 ABNs \citep{fukui2019attention} introduce an auxiliary attention module to find key regions of images to assist convolutional networks  in making predictions.
In contrast, our method directly modifies the cross-entropy loss rather than introducing an additional regularization term, so that robustness to context shifts is built in intrinsically.

An orthogonal approach is worst-group generalization \citep{arjovsky2020invariantriskminimization, chevalley2022invariant, sagawa2019distributionally}. Rather than optimizing for overall accuracy, this line of work partitions the dataset into groups defined by object-context combinations and seeks to balance performance across groups by improving the worst-performing group. 

\section{Motivation}
\label{sec:motivation}

In image classification tasks, deep neural networks make their predictions based on both the {\it object} and the {\it context}, i.e., the backgrounds in images.
It has been observed that irrelevant, or even inversely correlated, context can make a substantial contribution to the model predictions, lowering model's confidence on correct predictions, or even overturning the prediction into incorrectness with high confidence \citep{xiao2020noise,geirhos2020shortcut,Piekarek_2025_ICCV}.
This reliance renders the model less robust to context shifts, such as when a trained model is deployed in a different environment, and more vulnerable to adversarial manipulation; for instance, when an attacker deliberately replaces the context with one commonly associated with other classes.
In contrast, humans are able to distinguish between key features and spurious context; where deemed necessary, they can focus on the object region, neglecting the context, so as to accurately classify the relevant object in question.

As an illustration of this issue, consider the Waterbirds dataset \citep{sagawa2019distributionally}, which consists of two classes of birds: {\it waterbirds} and {\it landbirds}.
During training, most birds appear in their natural habitat: waterbirds are present by waterfronts, whereas landbirds are present surrounded by land.
However, during testing, a context shift may occur, where a landbird (or a waterbird) appears in a waterfront context (or a land context, respectively).
Neural networks often make incorrect predictions on images with context shift.
For example, with a fine-tuned ResNet-50, the context region on average accounts for $46\%$ of the model's prediction score compared to the influence from the object. Moreover, the model's accuracy drops from $98\%$ for normal images down to $41\%$ on images with context shift, which is even worse than the random guess accuracy of $50\%$.
Using HiResCAMs \citep{draelos2020use}, a member of the CAM (short for Class Activation Mapping) family that produces a heatmap highlighting the contribution from spatial locations of the image to the network output \citep{zhou2016learning}, we can visualize the context's contributions to the model prediction, see Figure \ref{fig:overview} for example. We see that when the image context is correlated with a different class, the misleading context makes large contributions to the prediction resulting in misclassifications with high confidence.  
This shows the pervasiveness of the context robustness concern, supported by broader statistics across other datasets in Table \ref{table:contribution-ratio}.

Our observation is that the neural networks, together with its optimization problem, do not possess an intrinsic mechanism to distinguish objects from context. 
To see this, consider a neural network $f$ (before the {\it softmax}), such as a ResNet-50, composed of a convolutional backbone $g$ and a bias-free linear classifier $h$: $f = h \circ g$. 
Given an input image $\rmX$, denote the feature embedding $g(\rmX) = \rmA \in \mathbb{R}^{d_0 \times d_1 \times d_2}$ and the final logit $f(\rmX) \in \mathbb{R}^K$, where $K$ is the number of classes, $d_0$ is the number of features (a.k.a.~channels) and $d_1$ and $d_2$ are the spatial dimensions of the final convolutional layer.
HiResCAMs, defined as follows,
\begin{equation}
	\hirescam{k} = \sum^{d_0}_{j=1} (\nabla_{\rmA_j} f_k) \odot \rmA_j, \quad \forall k \in [K], \label{eq:hi_res_cam}
\end{equation}
have the following relation with the network output logits \citep{draelos2020use}:
\begin{equation}\label{eq:logit-connection}
    f_k(\rmX) = \sum^{d_1,d_2}_{i,j=1} \left(\hirescam{k}(\rmX)\right)_{ij}, \quad \forall k \in [K].
\end{equation}
\begin{remark}\label{remark:gradcam_no_logit_connection}
    Other CAM methods, such as GradCAM \citep{selvaraju2017grad} and GradCAM++ \citep{Chattopadhay_2018}, do not possess the above relation with output logits.
\end{remark}
Let $H \in \mathbb{B}^{d_1 \times d_2}$ be the object indicator function, i.e., $H_{ij} = 1$ if the $(i,j)$-patch corresponds to the object and $H_{ij} = 0$ if the $(i,j)$-patch corresponds to the context. Then, the network's final output logits can equivalently be written as:
\begin{equation}\label{eq:biscet-hirescam}
    f_k(\rmX) = { \sum^{d_1,d_2}_{i,j=1}} ({\color{blue} H \odot \hirescam{k}(\rmX)})_{ij} + {\sum^{d_1,d_2}_{i,j=1}} ({\color{red} (1 - H) \odot \hirescam{k}(\rmX)})_{ij}, \quad \forall k \in [K],
\end{equation}
where the first and second terms correspond to contributions from the {\color{blue} object} and {\color{red} context} regions respectively. Here, we can see that the context contributes to the prediction in the same way as the object regions, and as seen in the demonstration in Figure \ref{fig:overview}, the network does not learn to suppress the context contributions. 

In this paper, we seek to alleviate this issue of sensitivity to context shift with a focus on convolutional neural networks. Specifically, we will utilize a variant of HiResCAMs to design a new loss function that is capable of suppressing context contributions, encouraging the models to make predictions by primarily focusing on the objects thus improving their robustness to context shift.

\section{From HiResCAMs to ContrastiveCAMs}
\label{sec:def-contrastive-cams}
In this section, we first present a fundamental issue of HiResCAMs in the multiclass setting, which degrades the faithfulness of feature attribution maps.
We then propose a variant, {\it ContrastiveCAMs}, which produces accurate, class-specific attribution maps even in the multiclass setting.

\paragraph{HiResCAMs Admit Spurious Shifts.}
A key drawback of HiResCAMs emerge as a consequence of applying {\it softmax} activation on the network's output {\it logits} $f$ to obtain the final predictions $\tilde{f}=\sigma(f)$, which are typically interpreted as {\it class probabilities}. 
\begin{proposition}
    [Redundancy in the {\it softmax} function]\label{prop:softmax-redundancy} 
	\begin{equation}
        \sigma(\rvx) = \sigma(\rvx + a \mathbf{1}_K), \qquad \forall\rvx \in \mathbb{R}^K,~a \in \mathbb{R}.
	\end{equation}
\end{proposition}
\begin{proof}
    The proof is deferred to Appendix \ref{sec:proofs}.
\end{proof}

The above proposition shows that, given a prediction $\tilde{f}$, the logit vector $f$ is not unique: if each logit is shifted by the same value, the final probability predictions $\tilde{f}$ remain unchanged.
Therefore, the value of each individual logit $f_k$ alone does not determine the final prediction.
This redundancy has also been observed recently by \citet{oh2025beyond, piland2026diffgradcam}, and discussed in the context of GradCAMs. 
Here, we show that this redundancy is inherited by HiResCAMs, and further pronounced in terms of degrees of freedom, via Eq.~(\ref{eq:logit-connection}). 

\begin{theorem}\label{thm:hirescam}
Let $\hirescam{} \in \mathbb{R}^{K \times d_1 \times d_2}$ be a 
HiResCAM. Given an arbitrary matrix $M\in \mathbb{R}^{d_1 \times d_2}$, define a new HiResCAM as:
\begin{equation}
    \overlinehirescam{k} := \hirescam{k} + M, \qquad \forall k \in [K].
\end{equation}
Then, the two HiResCAMs correspond to the same network prediction $\tilde{f}(\rmX)$:
\begin{equation}
    \tilde{f}(\rmX) = \sigma \left(\sum^{d_1, d_2}_{i, j = 1} \hirescam{:i j} \right) = \sigma \left(\sum^{d_1, d_2}_{i, j = 1} \overlinehirescam{:ij} \right).
\end{equation}
\end{theorem}

\begin{figure}[t]
    \centering
    \includegraphics[width=\linewidth]{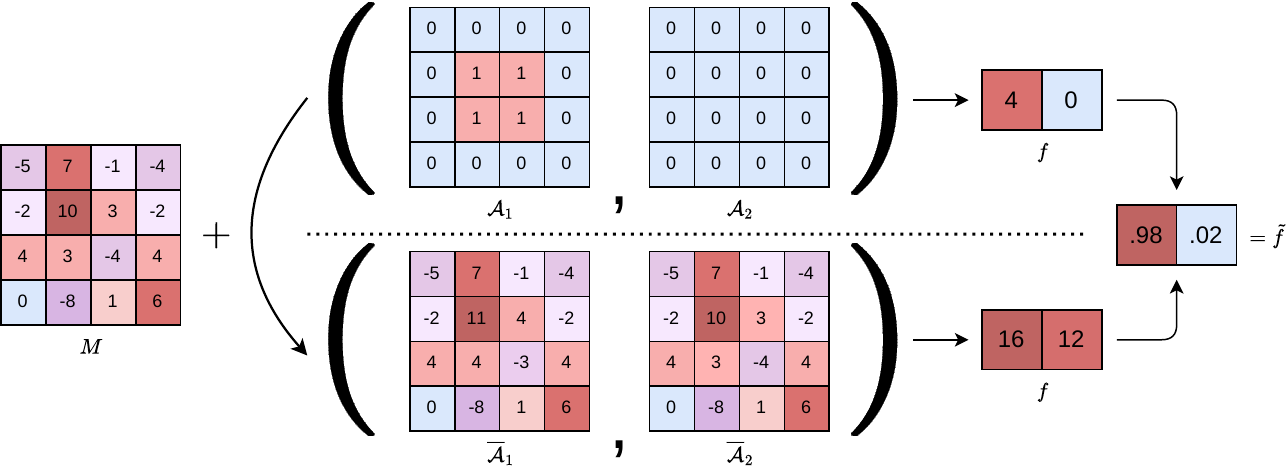}
    \caption{{\bf Illustration of Theorem \ref{thm:hirescam}.} Shifting $\hirescam{}$ by arbitrary matrix $M$ results in new explanations $\overlinehirescam{}$. Although the explanation changed significantly, it corresponds to an identical final prediction $\tilde{f}$.}
    \label{fig:hirescam-failure-mode}
\end{figure}

From this theorem, we can see that the feature attributions rendered by HiResCAMs can be shifted arbitrarily by a matrix $M \in \mathbb{R}^{d_1 \times d_2}$.
As a result, feature attributions can spuriously correspond to any given region without affecting the final prediction.
This is illustrated in Figure \ref{fig:hirescam-failure-mode}.
These additional degrees of freedom hinder the effectiveness of regularizing against object-context segmentation during training, since the explanation can be modified without influencing the overall prediction.

To address this issue, we propose to remove the redundant degrees in the feature attribution maps using a contrastive variant of HiResCAMs, termed {\it ContrastiveCAMs}.
\begin{definition}[$k$-ContrastiveCAMs]\label{def:contcams}
	Given a class of interest $k \in [K]$ and a set of HiResCAMs $\hirescam{}$ for a given image $\rmX$, $k$-ContrastiveCAMs   are defined as follows:
        \begin{equation}\label{eq:contrastive-cams}
		\contcam_{k} := \left\{ \contcamidx{k}{k'} : k' \in [K] \right\},
        \qquad \mathrm{where} \qquad
        \contcamidx{k}{k'} : = \hirescam{k} - \hirescam{k'}.
    \end{equation}
    The aggregated version of $k$-ContrastiveCAM is defined as
    \begin{align}
		\reconcam{k} := \frac{1}{K}\sum_{k'=1}^K \contcamidx{k}{k'} = \hirescam{k} - \frac{1}{K} \sum_{k'=1}^K \hirescam{k'}.
    \end{align}
\end{definition}
Here, $\contcamidx{k}{k'} $ is also referred as $(k, k')$-ContrastiveCAM.
Contrary to HiResCAMs $\hirescam{k}$, which focus on single-logit explanations, ContrastiveCAMs provide spatial feature attributions that explain why the model favors one class over a competing class, making them class-discriminative.

\begin{remark}\label{remark:hrc_sigmoid_contrastive}
For binary classification tasks where there is a single output logit and {\it sigmoid} applied to obtain the final predictions, HiResCAMs are intrinsically contrastive.
\end{remark}

\begin{remark} The idea of subtracting the average CAM has been applied in DiffGradCAMs \citep{piland2026diffgradcam}. However, DiffGradCAMs are built upon GradCAMs, whereas our ContrastiveCAMs are built upon HiResCAMs. With this distinction, both the following key properties hold for ContrastiveCAMs but not for DiffGradCAMs (c.f. Eq.(\ref{eq:logit-connection}) and Remark \ref{remark:gradcam_no_logit_connection}).
\end{remark}

\begin{proposition}[Uniqueness of ContrastiveCAMs]\label{prop:contcam-uniqueness}
Let $\hirescam{}$ and $\overlinehirescam{}$ be two HiResCAMs corresponding to probability predictions $\tilde{f}(\rmX) \in \mathbb{R}^K$ such that:
\begin{equation}
    \overlinehirescam{k} = \hirescam{k} + M, \qquad \forall k \in [K],
\end{equation}
with $M\in \mathbb{R}^{d_1 \times d_2}$ being an arbitrary matrix. Then, the ContrastiveCAMs $\contcam{}$  and $\overline{\contcam}$, derived from $\hirescam{}$ and $\overlinehirescam{}$ respectively, are the same:
\begin{equation}
    \contcam{} = \overline{\contcam},
    \qquad
    \mathrm{and}
    \qquad
    \reconcam{} = \overline{\contcam}^{ag}.
\end{equation}
\end{proposition}
Thus ContrastiveCAMs remove the extra degree of freedom, resulting in faithful feature attribution maps.
ContrastiveCAMs also have a close connection to the model's class probabilities.
Specifically, any {\it input-dependent} change to probability predictions $\tilde{f}$ (e.g., as a result of updating model parameters) is reflected precisely by a proportionate change to ContrastiveCAMs.
\begin{proposition}[Correctness of ContrastiveCAM Explanations]\label{prop:contcam-correctness}
Given an input $\rmX$ and a class index $k \in [K]$, the prediction $\tilde{f}_{k}(\rmX)$ is determined by the $k$-ContrastiveCAMs:
\begin{equation}\label{eq:relation-ccam-pred}
    \tilde{f}_{k}(\rmX) = \left(\sum^K_{k'=1}\exp\left(-\sum_{i,j=1}^{d_1,d_2} \left(\contcam_{(k, k')}\right)_{ij} \right)\right)^{-1}, \quad \forall k \in [K].
\end{equation}
Moreover, if for some $k\in [K], i \in [d_1], j \in [d_2]$, $\left(\contcam_{(k, k')}\right)_{ij} = 0, ~\forall k' \in [K]$,  then the $(i,j)$-patch does not influence the final prediction.
\end{proposition}

Comparing Eq.~(\ref{eq:relation-ccam-pred}) with Eq.~(\ref{eq:logit-connection}), one can easily find that, given any class index $k$, the corresponding HiResCAM only determines the output logit value $f_k$ but not the final prediction $\tilde{f}_k$; on the other hand, ContrastiveCAMs directly determines the final prediction $\tilde{f}_k$, which makes ContrastiveCAMs a more faithful tool to explain the model's prediction process. Moreover, the above proposition also shows that, iff the ContrastiveCAMs are zero in a certain spatial region, the contribution from this region of the image to the prediction is indeed zero.
HiResCAMs do not share this property.

\section{Learning with Context-Regularized Cross-Entropy Loss}
We now introduce a learning framework designed to be robust to context shifts.
In particular, we propose a novel loss function, called {\it Context-Regularized Cross-Entropy} (CR-CE), which inhibits the context from contributing to the model prediction.
This regularization is guided using the feature attributions produced by ContrastiveCAMs.

To effectively regularize against context contributions during training, prior knowledge of the contextual regions in the input images is generally required. Such information can be obtained in two main ways: (i) through segmentation annotations provided by the dataset (e.g., PASCAL VOC \citep{hariharan2011semantic}, MS-COCO \citep{lin2014microsoft}), or (ii) by leveraging pretrained object detection or segmentation models (e.g., Segment Anything Model \citep{carion2026sam}, Masked AutoEncoders \citep{he2022masked}) to estimate bounding boxes or object masks (see experiments on ImageNet, Spawrious datasets in Section \ref{sec:experiments}).
In practice, the first option is often unavailable or prohibitively expensive. We adopt the second approach which is more accessible, as pretrained models can provide approximate context localization with minimal effort. Note that, our framework does not require precise object segmentation or localization, and rough estimates are often sufficient to effectively regularize contextual influence.
In Appendix \ref{sec:crce-tolerance-noisy-masks}, we show numerical evidence for the robustness of our framework to noisy object-context segmentation.

Given an image $\rmX$, denote $H\in \mathbb{B}^{d_1\times d_2}$ as the binary mask that distinguishes object regions from context. Specifically, for each $(i,j)$-patch where $i\in [d_1], j\in[d_2]$, $H_{ij}=1$ if the patch corresponds to the object, and $H_{ij}=0$ if it corresponds to the context. We define CR-CE as follows:
\begin{definition}[Context-Regularized Cross-Entropy Loss]\label{def:crce}
Given a training dataset $\mathcal{D}$ and convolutional network $f$, the CR-CE loss is defined as:
    \begin{align}
	    &\mathcal{L}_{\mathrm{CRCE}}(\mathcal{D}) = \mathbb{E}_{(\rmX,\rvy,H)\sim \mathcal{D}}[\ell(\rmX,\rvy, H)], ~~\textrm{with}\nonumber\\
        \ell(\rmX,\rvy, H) &= \log \left(\sum^K_{k=1} \exp \left(\sum_{i,j=1}^{d_1, d_2} -H_{ij} \left(\contcamidx{k_t}{k}\right)_{ij} + \underbrace{(1 - H_{ij}) \left|\right(\contcamidx{k_t}{k}\left)_{ij}\right|}_{\text{context regularization}} \right)\right).
	\end{align}
where $k_t$ is the index of the ground-truth class. For simplicity of notation, we do not explicitly write the dependence of $H$ and ContrastiveCAMs $\contcam$ on the input image $\rmX$.
\end{definition}

The context regularization term in the CR-CE loss regularizes the absolute values of the ContrastiveCAMs $|(\contcamidx{k_t}{k})_{ij}|$ at context regions, driving them towards zero, thus minimizing the influence of context in the final prediction $\tilde{f}$ (c.f. Proposition \ref{prop:contcam-correctness}). 
We also extend context regularization to binary classification settings; we defer this discussion to Appendix \ref{sec:crbce}.

\paragraph{Comparison with standard cross-entropy loss.} The only difference between the standard cross-entropy (CE) loss and CR-CE lies in the context regularization term. In particular, if the absolute values of the ContrastiveCAM, $|(\contcamidx{k_t}{k})_{ij}|$, are replaced with their negated values, $-(\contcamidx{k_t}{k})_{ij}$, the CR-CE loss reduces to the standard cross-entropy loss, noting that cross-entropy loss on one sample $\ell_{CE}(\rmX,\rvy) = -\log \tilde{f}_{k_t}$ as well as the relation in Proposition \ref{prop:contcam-correctness}.

Despite this seemingly minor modification, the effect on model behavior is substantial. The standard cross-entropy loss implicitly encourages the context to cooperate with the  object regions in making predictions. While this is beneficial when the context is consistent with the object, it can lead to an over-reliance on contextual cues and thus reduce robustness under context shift. In contrast, CR-CE explicitly penalizes excessive reliance on context, encouraging the model to base its predictions primarily on the object itself. As a result, when the context is altered or adversarially manipulated, models trained with CR-CE exhibit improved robustness.

Further, for models using a linear classifier head (e.g., ResNet \citep{he2016deep}, MobileNet \citep{howard2019searching}, EfficientNet \citep{tan2019efficientnet}), ContrastiveCAM can be computed in closed form, see Eqs.~(\ref{eq:hi_res_cam}) and (\ref{eq:contrastive-cams}); this results in a minor computational overhead when CR-CE is used in place of cross-entropy.

\section{Experimental Results}
\label{sec:experiments}

In this section, we numerically evaluate the effectiveness of training with the proposed CR-CE loss function in both promoting object-focused learning and improving robustness to context shifts.

\subsection{Datasets and Experimental Setup}
\paragraph{Datasets.}
We evaluate our method using the following four datasets:

\begin{itemize}[leftmargin=*]
    \item {\bf Waterbirds-100} dataset \citep{sagawa2019distributionally} consists of images of {\it waterbirds} and {\it landbirds}.
    In the training set, each class of bird appears in the context of their natural habitat--{\it waterfront} and {\it land}, respectively.
    However, in the test set, context shifts may occur, with birds appearing in the opposite environment. 
    \item {\bf Spawrious} dataset \citep{lynch2023spawrious} consists of AI-generated images of four classes of dog breeds appearing in various contexts. The test set contains combinations of dog breed and context that are not seen during training, thereby assessing models' robustness to context shift.
    \item {\bf ImageNet} \citep{deng2009imagenet} with {\bf ImageNet-BG} \citep{Piekarek_2025_ICCV} as test set. We train the model using ImageNet-1k and test it using ImageNet-BG, the latter of which features images with manually replaced backgrounds, designed to decouple object recognition from context cues.
    \item {\bf Hard-ImageNet} \citep{moayeri2022hard} is a subset of ImageNet, consisting of $15$ selected classes of images that have high context influence.
    It is used to investigate the context's influence on the classification by comparing the model outputs with and without the objects of the images.
\end{itemize}

For all the datasets above, context shifts only occur in the test set, which is not accessed during training.
Ground-truth object-context masks are available for the training sets of Waterbirds-100 and Hard-ImageNet.
For Spawrious and ImageNet where object-context segmentation is not available, 
we use existing segmentation models such as SAM3 \citep{carion2026sam} and Masked AutoEncoders (MAE) \citep{he2022masked, gao2021luss} to generate these masks.
Our approach allows for imperfect segmentation, enabling the effective use of off-the-shelf vision models when ground truth masks are unavailable.

\paragraph{Experimental Setup.}
We follow the methodology in the benchmark datasets \citep{sagawa2019distributionally, lynch2023spawrious, Piekarek_2025_ICCV, moayeri2022hard} and fine-tune ResNet-50 models pre-trained on ImageNet-1k.
In our experiments, we also evaluate a modified ResNet architecture where the final downsampling convolution layer's stride is set to $(1, 1)$, thus doubling the resolution of ContrastiveCAMs, i.e., $d_1 = d_2 = 14$.
The higher-resolution attribution maps provide more precise localization of object regions.
More experiments with different architectures can be found in Appendix \ref{sec:crce-effective-across-archs}.
We compare our proposed CR-CE framework with the following prior approaches:
\begin{itemize}[leftmargin=*]
    \item The standard training procedure with cross-entropy loss (CE), which does not implement any context-shift robustness technique. It serves as a baseline for comparison.
    \item Explanation-guided methods, which improve context-shift robustness by guiding models' feature attributions, i.e., the heatmaps, towards the target object region. Our proposed method falls in this category. 
    We compare with existing methods: attention branch networks (ABN) \citep{fukui2019attention}, GALS \citep{petryk2022gals}, and training with energy-based loss (EBL) \citep{rao2023studying}.
    \item Other methods, including Context Masking training \citep{aniraj2023masking} and Causal Feature Alignment \citep{venkataramani2024causal}. Both methods use object-context masks to transform the input/hidden features 
    in order to discourage context reliance. Particularly, Context Masking only uses the object regions of images for training by masking out all context regions. 

\end{itemize}

For fairness of comparison, we use identical object-context masks across approaches.
For all our metrics, we report the mean and standard deviation computed over 3 independent runs.
Details of the experimental setup are described in Appendix \ref{sec:training-detail}.

\begin{figure}[t]
    \centering
    \includegraphics[width=\linewidth]{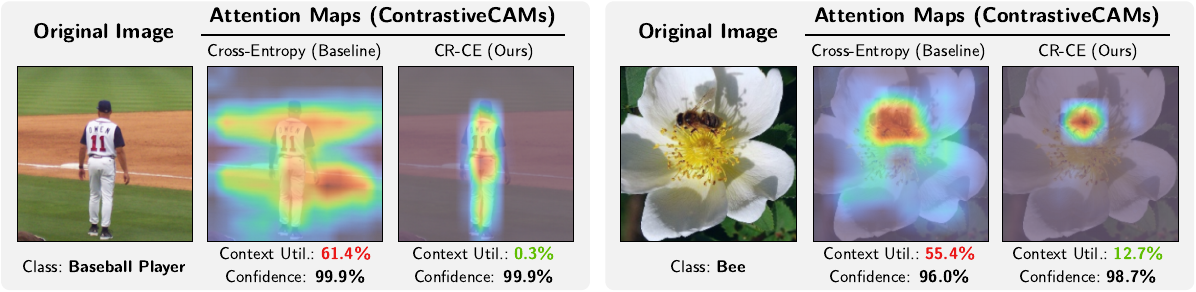}
    \caption{{\bf Qualitative Evaluation.}
    Models trained using cross-entropy rely heavily on context: in the Hard-ImageNet (left) and ImageNet-BG (right) examples shown above, the {\it baseball player} and {\it bee} are predicted using the {\it playing field} ($61.4\%$ of the prediction) and the {\it flower} ($55.4\%$ of the prediction) respectively.
    On the other hand, models trained using CR-CE leverage object-specific regions and localize the target without the use of object-context segmentation during inference.
    }
    \label{fig:crce-imagenetbg}
\end{figure}

\subsection{CR-CE Suppresses Context Contributions}

First, we evaluate the proportion of context contributions using the following metric: 
\begin{equation}
    R(\reconcam{}, H) = \frac{\sum^{d_1, d_2}_{i,j=1} |(\reconcam{k_t})_{ij}| (1-H_{ij})}{\sum_{i,j=1}^{d_1, d_2} |(\reconcam{k_t})_{ij}|},
\end{equation}
where $k_t \in [K]$ is the target class index.
This metric represents the portion of the absolute value of the {\it ContrastiveCAM} heatmap from the context region.
A comparison of the proposed CR-CE training framework with the standard CE training is shown in Table \ref{table:contribution-ratio}, with Figure \ref{fig:crce-imagenetbg} and Figure \ref{fig:imagenet_hard_bg_examples} presenting examples.
Comparisons with the other methods assessed in this work are included in Table \ref{table:contribution-ratio-appendix}. 
We observe that training with CR-CE effectively guides ContrastiveCAMs onto the object regions and reduces the model’s reliance on the context region.

Additional evaluation of the effectiveness of CR-CE in suppressing context contributions using the Hard-ImageNet evaluation suite can be found in 
Appendix \ref{sec:hardinet_evals}.

\subsection{Reduced Context-Contributions Improve Robustness to Context Shifts}
\label{sec:crce-improves-robustness}

\begin{table}[t]

\caption{{\bf Context Contribution Ratios.}
On the test set of each context-shift robustness benchmark we consider, CR-CE trained models primarily rely on object-specific regions to determine predictions.}
\label{table:contribution-ratio}
\centering

\begin{adjustbox}{max width=\textwidth}
\begin{threeparttable}
\begin{tabular}{cccccc}
\toprule
\multirow{2}[3]{*}{\textbf{\begin{tabular}[c]{@{}c@{}}CAM \\ Size\end{tabular}}} &
\multirow{2}[3]{*}{\textbf{Approach}} & 
\multicolumn{4}{c}{\textbf{Context Contribution Ratio $R\%$ $(\downarrow)$}}                                                    \\
\cmidrule{3-6}
      &                           &  Waterbirds-100  & Spawrious\tnote{\textdaggerdbl} &  ImageNet-BG & Hard-ImageNet  \\
      \midrule
      \multirow{2}{*}{$7 \times 7$}  &  CE (Baseline)  & $46.08${\tiny $\pm 0.43$} &	$44.79${\tiny $\pm 0.3$}	& $31.83${\tiny $\pm 0$}	& $68.53${\tiny $\pm 0.04$} \\
      & CR-CE (Ours) & $\mathbf{12.86}${\tiny $\pm 0.09$} &	$\mathbf{14.89}${\tiny $\pm 0.2$}	& $\mathbf{14.85}${\tiny $\pm 0$}	& $\mathbf{35.85}${\tiny $\pm 0.26$} \\
      \midrule
      \multirow{2}{*}{$14 \times 14$}  & CE (Baseline) & $58.75${\tiny $\pm 0.12$} &	$47.32${\tiny $\pm 0.09$}	& $28.27${\tiny $\pm 0.27$}	& $66.5${\tiny $\pm 0.07$} \\
      & CR-CE (Ours) & $\mathbf{9.54}${\tiny $\pm 0.28$}  & $\mathbf{10.98}${\tiny $\pm 0.12$}	& $\mathbf{9.32}${\tiny $\pm 0.06$}	& $\mathbf{31.12}${\tiny $\pm 0.1$} \\
      \bottomrule
\end{tabular}

\begin{tablenotes}
    \small
    \item [\textdagger] No std.~dev.~is reported; we use pre-trained weights from \texttt{torchvision} without modification.
    \item [\textdaggerdbl] We report metrics on the hardest subtask and difficulty (M2M-hard) within the Spawrious benchmark.
\end{tablenotes}
\end{threeparttable}
\end{adjustbox}
\end{table}
\begin{table}[t]
\caption{
{\bf Context-Shift Robustness on Waterbirds-100, ImageNet, and Spawrious datasets.}
Accuracy across various benchmarks, with the best context-shifted performance highlighted in {\bf bold} and the highest in-distribution performance \underline{underlined}.
Models trained with CR-CE generalize effectively to novel environments at a minor cost to in-distribution performance.
}
\label{table:results_overview}
\centering
\begin{adjustbox}{max width=\textwidth}
\begin{threeparttable}[h]
\begin{tabular}{ccccccccc}
\toprule
\multirow{2}[3]{*}{\textbf{\begin{tabular}[c]{@{}c@{}}CAM \\ Size\end{tabular}}} &
\multirow{2}[3]{*}{\bf Approach} &
  \multicolumn{2}{c}{\bf Waterbirds-100} &
  \multicolumn{2}{c}{\bf ImageNet/ImageNet-BG} &
  \multicolumn{3}{c}{\bf Spawrious\tnote{\textdaggerdbl}} \\
  \cmidrule(lr){3-4}
  \cmidrule(lr){5-6}
  \cmidrule(lr){7-9}
 &
  \multicolumn{1}{c}{} &
  \multicolumn{1}{c}{No Shift $(\uparrow)$} &
  \multicolumn{1}{c}{Shifted $(\uparrow)$} &
  \multicolumn{1}{c}{No Shift $(\uparrow)$} &
  \multicolumn{1}{c}{Shifted $(\uparrow)$} &
  \multicolumn{1}{c}{Easy $(\uparrow)$} &
  \multicolumn{1}{c}{Medium $(\uparrow)$} &
  \multicolumn{1}{c}{Hard $(\uparrow)$} \\
\midrule
\multirow{6}{*}{$7 \times 7$} & CE (Baseline) & $\underline{98.66}${\tiny $\pm 0.10$} & $41.74${\tiny $\pm 4.15$} & $\underline{82.92}$\tnote{\textdagger} & $74.83$\tnote{\textdagger} & $90.86${\tiny $\pm 0.54$} & $79.46${\tiny $\pm 0.48$} & $81.52${\tiny $\pm 1.03$} \\
& GALS & $97.54${\tiny $\pm 1.01$} & $54.45${\tiny $\pm 6.75$} & $82.68${\tiny $\pm 0.06$} & $73.82${\tiny $\pm 0.10$} & $93.47${\tiny $\pm 0.36$} & $84.58${\tiny $\pm 0.88$} & $78.74${\tiny $\pm 0.97$} \\
& EBL & $96.35${\tiny $\pm 0.49$} & $51.31${\tiny $\pm 4.70$} & $82.88${\tiny $\pm 0.08$} & $74.79${\tiny $\pm 0.01$} & $92.43${\tiny $\pm 1.31$} & $87.23${\tiny $\pm 0.84$} & $81.21${\tiny $\pm 0.74$} \\
& CFA & $93.43${\tiny $\pm 0.13$} & $72.71${\tiny $\pm 0.64$} & $82.71${\tiny $\pm 0.03$} & $73.85${\tiny $\pm 0.09$} & $89.14${\tiny $\pm 0.89$} & $85.68${\tiny $\pm 0.68$} & $81.77${\tiny $\pm 1.73$} \\
& Context Masking & $91.16${\tiny $\pm 3.18$} & $61.67${\tiny $\pm 6.17$} & $80.21${\tiny $\pm 0.07$} & $72.87${\tiny $\pm 0.10$} & $92.08${\tiny $\pm 1.04$} & $85.03${\tiny $\pm 2.37$} & $78.46${\tiny $\pm 1.12$} \\
& {\bf CR-CE (Ours)} & $94.38${\tiny $\pm 1.21$} & $\mathbf{73.30}${\tiny $\pm 4.43$} & $81.85${\tiny $\pm 0.03$} & $\mathbf{75.75}${\tiny $\pm 0.08$} & $\mathbf{95.55}${\tiny $\pm 0.16$} & $\mathbf{91.25}${\tiny $\pm 0.18$} & $\mathbf{87.49}${\tiny $\pm 0.14$} \\
\midrule
\multirow{7}{*}{$14 \times 14$} & CE (Baseline) & $\underline{98.48}${\tiny $\pm 0.09$} & $38.40${\tiny $\pm 1.79$} & $82.33${\tiny $\pm 0.10$} & $74.72${\tiny $\pm 0.06$} & $88.98${\tiny $\pm 0.14$} & $73.38${\tiny $\pm 0.87$} & $79.76${\tiny $\pm 1.01$} \\
& ABN & $97.39${\tiny $\pm 0.55$} & $51.87${\tiny $\pm 6.14$} & $79.91${\tiny $\pm 0.06$} & $69.07${\tiny $\pm 0.15$} & $89.65${\tiny $\pm 0.44$} & $72.81${\tiny $\pm 1.32$} & $63.96${\tiny $\pm 6.54$} \\
& GALS & $97.39${\tiny $\pm 0.28$} & $63.21${\tiny $\pm 4.90$} & $82.43${\tiny $\pm 0.04$} & $75.21${\tiny $\pm 0.02$} & $95.03${\tiny $\pm 1.35$} & $87.63${\tiny $\pm 3.80$} & $84.08${\tiny $\pm 1.87$} \\
&  EBL & $96.33${\tiny $\pm 0.95$} & $51.49${\tiny $\pm 5.31$} & $\underline{82.72}${\tiny $\pm 0.09$} & $75.79${\tiny $\pm 0.20$} & $93.46${\tiny $\pm 0.73$} & $89.05${\tiny $\pm 0.47$} & $79.36${\tiny $\pm 2.83$} \\
& CFA & $93.09${\tiny $\pm 0.74$} & $77.51${\tiny $\pm 3.89$} & $79.54${\tiny $\pm 0.09$} & $73.48${\tiny $\pm 0.03$} & $87.22${\tiny $\pm 0.78$} & $85.68${\tiny $\pm 0.68$} & $75.88${\tiny $\pm 1.12$} \\
& Context Masking & $92.27${\tiny $\pm 1.18$} & $64.39${\tiny $\pm 1.79$} & $80.70${\tiny $\pm 0.07$} & $75.49${\tiny $\pm 0.10$} & $90.96${\tiny $\pm 0.24$} & $85.70${\tiny $\pm 2.99$} & $80.80${\tiny $\pm 0.71$} \\
& {\bf CR-CE (Ours)} & $94.36${\tiny $\pm 0.12$} & $\mathbf{79.13}${\tiny $\pm 0.77$} & $81.25${\tiny $\pm 0.03$} & $\mathbf{76.84}${\tiny $\pm 0.12$} & $\mathbf{96.47}${\tiny $\pm 0.06$} & $\mathbf{92.41}${\tiny $\pm 0.29$} & $\mathbf{88.66}${\tiny $\pm 0.52$} \\
\bottomrule
\end{tabular}%
\begin{tablenotes}
\small
    \item [\textdagger] No std.~dev.~is reported; we use pre-trained weights from \texttt{torchvision} without modification.
    \item [\textdaggerdbl] The reported metrics correspond to the many-to-many spurious correlation setting; the complete benchmark is assessed in Appendix \ref{sec:spawrious_all_configuration_results}.
\end{tablenotes}
\end{threeparttable}
\end{adjustbox}
\vspace{-15pt}
\end{table}

Next, we demonstrate the improved performance of CR-CE trained models under context shift by comparing them against the standard cross-entropy training, existing explanation-guided learning approaches \citep{petryk2022gals, fukui2019attention,rao2023studying} and other methods  \citep{venkataramani2024causal, aniraj2023masking}.
We report the numerical results
in Table \ref{table:results_overview}.

We observe that across all datasets, the CR-CE training framework consistently improves the robustness to context shift and further outperforms models trained using cross-entropy loss and the other compared methods.
This demonstrates that the proposed method successfully suppresses contribution from irrelevant or misleading context, yielding correct and robust model predictions.
We also note that, when the context is relevant (as in the setting without context shift), there is a slight degradation in the prediction accuracy across tasks.
This is expected because when the test dataset and training dataset have the same set of object-context combinations, the model can often infer the correct class with the help of the context.
Indeed, such correlations are often not desirable if context-shift robustness is the target.

Additionally, we observe that some existing explanation-guided methods may underperform the standard cross-entropy baseline on complex classification tasks,  such as ImageNet, while outperforming it on simpler tasks such as the Waterbirds-100 dataset.
We hypothesize that this behavior is largely due to their reliance on classical feature attribution methods (e.g. Grad-CAM and HiResCAM), which suffer from the unfaithfulness issue theoretically described in Theorem \ref{thm:hirescam}. In the binary classification setting of Waterbirds-100, HiResCAM coincides with our ContrastiveCAMs (see Remark \ref{remark:hrc_sigmoid_contrastive}) and does not exhibit this limitation. 
More discussion of explanation-guided methods on this matter can be found in 
 Appendix \ref{sec:egl-comparison}.

Notably, we emphasize that our approach does not use object-context segmentation during inference; the promising performance of our approach suggests that the CR-CE trained models have the ability to distinguish objects from context while making predictions.

Finally, we also note that the generalization gap between in-distribution and context-shifted images shrinks as explanation granularity is increased (comparing the $7\times 7$ and $14\times 14$ CAM resolutions).

\section{Conclusion}
\label{sec:discussion}
In this work, we advance explanation faithfulness and context robustness in convolutional networks.
We introduce {\it ContrastiveCAMs}, a feature attribution method that addresses limitations of existing CAM methods and has a strong connection to model predictions.
We further propose a new loss function, {\it Context-Regularized Cross-Entropy} (CR-CE), which penalizes contributions from context regions, thereby improving models' robustness to context shifts. 
Our experimental evaluations show that our method effectively attracts the
feature attributions towards the object region, and enhances context-shift robustness.

\paragraph{Limitations.} We identify two limitations of this work. First, the method does not naturally extend to transformer architectures, such as Vision Transformers (ViTs) \citep{alexey2020image}. This is because CAM-based spatial feature attribution methods rely on the local structure induced by convolutional layers, whereas attention mechanisms \citep{vaswani2017attention} mix patch tokens in a way that loses explicit spatial locality.
Second, the proposed CR-CE training requires additional object-context segmentation information (albeit
only during training), which requires additional cost to obtain in practice. We note that existing
explanation-guided methods also rely on this information.

\section*{Acknowledgment}

This research used both the DeltaAI advanced computing and data resource, which is supported by the
National Science Foundation (award OAC 2320345) and the State of Illinois, and the Delta advanced
computing and data resource which is supported by the National Science Foundation (award OAC 2005572)
and the State of Illinois. Delta and DeltaAI are joint efforts of the University of Illinois Urbana-Champaign
and its National Center for Supercomputing Applications.

\bibliographystyle{plainnat}
\bibliography{references}

\newpage
\appendix
\onecolumn

\section{Proofs}\label{sec:proofs}

{\bf Proposition \ref{prop:softmax-redundancy}} {(Redundancy in the {\it softmax} function)}{\bf .}
\begin{equation}
    \sigma(\rvx) = \sigma(\rvx + a \mathbf{1}_K), \qquad \forall\rvx \in \mathbb{R}^K,~a \in \mathbb{R}.
\end{equation}
\begin{proof}
    \begin{equation}
        \sigma(\rvx + a \mathbf{1}_K)
        = \frac{1}{\sum_{k=1}^K e^{\rvx_k + a}}(e^{\rvx_1 + a},e^{\rvx_2 + a},\cdots,e^{\rvx_K + a})
        = 
            \frac{1}{\sum_{k=1}^K e^{\rvx_k}}(e^{\rvx_1},e^{\rvx_2},\cdots,e^{\rvx_K})
        = \sigma(\rvx).
    \end{equation}
\end{proof}

\begin{em}
    {\bf Theorem \ref{thm:hirescam}.} Let $\hirescam{} \in \mathbb{R}^{K \times d_1 \times d_2}$ be a HiResCAM.
    Given an arbitrary matrix $M\in \mathbb{R}^{d_1 \times d_2}$, define a new HiResCAM as:
    \begin{equation}
        \overlinehirescam{k} := \hirescam{k} + M, \qquad \forall k \in [K].
    \end{equation}
    Then, the two HiResCAMs correspond to the same network prediction $\tilde{f}(\rmX)$:
    \begin{equation}
        \tilde{f}(\rmX) = \sigma \left(\sum^{d_1, d_2}_{i, j = 1} \hirescam{:ij} \right) = \sigma \left(\sum^{d_1, d_2}_{i, j = 1} \overlinehirescam{:ij} \right).
    \end{equation}
\end{em}
\begin{proof}
The proof for this statement follows from the relation between the two HiResCAMs.
For a given shift $M$, the predictions from the corresponding HiResCAM $\overlinehirescam{}$ can be expressed as:
\begin{align}
    \sigma \left(\sum^{d_1, d_2}_{i, j = 1} \overlinehirescam{:ij} \right) 
    &= \sigma \left(\sum^{d_1, d_2}_{i, j = 1} (\hirescam{:ij} + M_{ij}) \right).
    \intertext{Let $a = \sum^{d_1, d_2}_{i,j=1} M_{ij}$. We can equivalently write the above statement as:}
    &= \sigma \left(\sum^{d_1, d_2}_{i, j = 1} \hirescam{:ij} + a \mathbf{1}_K \right).
    \intertext{Applying Proposition \ref{prop:softmax-redundancy}, we have:}
    \sigma \left(\sum^{d_1, d_2}_{i, j = 1} \overlinehirescam{:ij} \right) 
    &= \sigma \left(\sum^{d_1, d_2}_{i, j = 1} \hirescam{:ij} \right) = \tilde{f}(\rmX).
\end{align}
Proving the desired statement.
\end{proof}

{\bf Proposition \ref{prop:contcam-uniqueness}} (Uniqueness of ContrastiveCAMs){\bf .}
{\em Let $\hirescam{}$ and $\overlinehirescam{}$ be two HiResCAMs corresponding to probability predictions $\tilde{f}(\rmX) \in \mathbb{R}^K$ such that:
\begin{equation}
    \overlinehirescam{k} = \hirescam{k} + M, \qquad \forall k \in [K],
\end{equation}
with $M\in \mathbb{R}^{d_1 \times d_2}$ being an arbitrary matrix. Then, the ContrastiveCAMs $\contcam{}$ and $\overline{\contcam}$, derived from $\hirescam{}$ and $\overlinehirescam{}$ respectively, are the same:
\begin{equation}
    \contcam{} = \overline{\contcam},
    \qquad
    \mathrm{and}
    \qquad
    \reconcam{} = \overline{\contcam}^{ag}.
\end{equation}} 
\begin{proof}
    First, consider $k$-ContrastiveCAMs. We have that:
    \begin{align}
        \overline{\contcam}_k &= \{ \overline{\contcam}_{(k, k')} : k' \in [K] \}  \\
        &= \{ \hirescam{k} + M - \hirescam{k'} - M : k' \in [K] \}  \\
        &= \{ \hirescam{k} - \hirescam{k'} : k' \in [K] \}  \\
        &= \{ \contcamidx{k}{k'} : k' \in [K] \} = \contcam_k,
        \quad
        \forall k \in [K].\label{eq:contcam-invariant-m}
    \end{align}
    Similarly, in the case of aggregated $k$-ContrastiveCAMs, we have:
    \begin{align}
        \overline{\contcam}^{ag}_k &= \overlinehirescam{k} - \frac{1}{K} \sum^K_{k'=1} \overlinehirescam{k'}  \\
        &= \hirescam{k} + M - \frac{1}{K} \sum^K_{k'=1} (\hirescam{k'} + M)  \\
        &= \hirescam{k} + M - \frac{K \cdot M}{K} - \frac{1}{K} \sum^K_{k'=1} \hirescam{k'}  \\
        &= \hirescam{k} - \frac{1}{K} \sum^K_{k'=1} \hirescam{k'} = \reconcam{k},
        \quad
        \forall k \in [K].\label{eq:agg-contcam-invariant-m}
    \end{align}
    Eqs.~(\ref{eq:contcam-invariant-m}) and (\ref{eq:agg-contcam-invariant-m}) together prove the desired equalities.
\end{proof}

{\bf Proposition \ref{prop:contcam-correctness}}(Correctness of ContrastiveCAM Explanations){\bf .}
\begin{em}
Given an input $\rmX$ and a class index $k \in [K]$, the prediction $\tilde{f}_{k}(\rmX)$ is determined by the $k$-ContrastiveCAMs:
\begin{equation}
    \tilde{f}_{k}(\rmX) = \left(\sum^K_{k'=1}\exp\left(-\sum_{i,j=1}^{d_1,d_2} \left(\contcam_{(k, k')}\right)_{ij} \right)\right)^{-1}, \quad \forall k \in [K].
\end{equation}
Moreover, if for some $k\in [K], i \in [d_1], j \in [d_2]$, $\left(\contcam_{(k, k')}\right)_{ij} = 0, ~\forall k' \in [K]$,  then the $(i,j)$-patch does not influence the final prediction.
\end{em}
\begin{proof}
    For a class index $k \in [K]$, the prediction probability given a logit vector $f$ is expressed as:
	\begin{equation}
		  \tilde{f}_{k} = \sigma_{k}(f) = \frac{e^{f_{k}}}{\sum^K_{k'=1} e^{f_{k'}}}.
	\end{equation}

	We define the logit vector in terms of the elementwise difference to a candidate class $k$:
	\begin{gather}
		\rvd := f - f_k \implies f = f_k + \rvd.
	\end{gather}
	Based on this definition, class probabilities can equivalently be computed as:
	\begin{equation}
            \tilde{f}_{k} = \frac{e^{f_{k}}}{\sum_{k'=1}^K e^{f_{k'}}} = \frac{e^{f_{k}}}{\sum_{k'=1}^K e^{f_{k} +\rvd_{k'}}} = \frac{e^{f_{k}}}{e^{f_{k}} \sum_{k'=1}^K e^{\rvd_{k'}}} = \frac{1}{\sum_{k'=1}^K e^{\rvd_{k'}}} \label{eq:recontext-softmax}.
	\end{equation}
	This re-contextualizes softmax as a direct function of the differences between class logits. We can further deconstruct the difference by logit values:
    \begin{equation}
	\rvd_{k'} = f_{k'} - f_{k} = \sum^{d_1,d_2}_{i,j=1} \hirescam{k'ij} - \sum^{d_1,d_2}_{i,j=1} \hirescam{kij}, \quad \forall k, k' \in [K].
    \end{equation}
      Applying Definition \ref{def:contcams}, we have:
    \begin{equation}
    	\rvd_{k'} = \sum^{d_1,d_2}_{i,j=1} (\hirescam{k'} - \hirescam{k})_{ij} = -\sum^{d_1, d_2}_{i,j=1} \left(\contcamidx{k}{k'}\right)_{ij}, \quad \forall k, k' \in [K]. \label{eq:diff-as-contcam}
    \end{equation}
    Substituting $\rvd_{k'}$ from Eq.~(\ref{eq:diff-as-contcam}) into Eq.~(\ref{eq:recontext-softmax}), we have:
    \begin{align}
        \tilde{f}_{k}(\rmX) &= \frac{1}{\sum^K_{k'=1} \exp\left(-\sum_{i,j=1}^{d_1, d_2} \left(\contcamidx{k}{k'}\right)_{ij} \right)}  \\
        &= \left( \sum^K_{k'=1} \exp\left(-\sum_{i,j=1}^{d_1, d_2} \left(\contcamidx{k}{k'}\right)_{ij} \right) \right)^{-1}, \quad \forall k \in [K].\label{eq:cc-correctness-eq}
    \end{align}
    We can thus compute class probabilities as a direct function of ContrastiveCAMs.

    Next, we show that zero-patches do not influence predictions.
    For a given class $k \in [K]$, denote the set of zero-patches as $\mathcal{Z}_k$, and its complement as $\mathcal{Z}'_k$:
    \begin{equation}
        \mathcal{Z}_k = \left\{ (i, j) \in [d_1] \times [d_2]:  \left( \contcamidx{k}{k'} \right)_{ij} = 0,~\forall k' \in [K] \right\},
        \quad
        \mathrm{and}
        \quad
        \mathcal{Z}_k' = ([d_1] \times [d_2])\setminus \mathcal{Z}_k.
    \end{equation}
    Applying this partition to Eq.~(\ref{eq:cc-correctness-eq}), we have that:
    \begin{align}
        \tilde{f}_{k}(\rmX)
        &= \left( \sum^K_{k'=1} \exp\left(-\sum_{i,j=1}^{d_1, d_2} \left(\contcamidx{k}{k'}\right)_{ij} \right) \right)^{-1} \\
        &= \left( \sum^K_{k'=1} \exp\left(
            -\sum_{i,j \in \mathcal{Z}_k'} \left(\contcamidx{k}{k'}\right)_{ij}
            -\sum_{i,j \in \mathcal{Z}_k} \left(\contcamidx{k}{k'}\right)_{ij}
        \right) \right)^{-1} \\
        &= \left( \sum^K_{k'=1} \exp\left(
            -\sum_{i,j \in \mathcal{Z}_k'} \left(\contcamidx{k}{k'}\right)_{ij}
            -\sum_{i,j \in \mathcal{Z}_k} 0
        \right) \right)^{-1} \\
        &= \left( \sum^K_{k'=1} \exp\left(
            -\sum_{i,j \in \mathcal{Z}_k'} \left(\contcamidx{k}{k'}\right)_{ij}
        \right) \right)^{-1}.\label{eq:zero_patches_specific_k}
    \end{align}
    Note that, for some $k\in[K]$, $\left( \contcamidx{k}{k'} \right)_{ij} = 0$ for all $k'\in[K]$ is equivalent to $\left( \hirescam{k}\right)_{ij} = \left( \hirescam{k'}\right)_{ij}$ for all $k'\in[K]$, which further implies that $\left(\contcamidx{k}{k'}\right)_{ij} = 0$ for all $k,k'\in [K]$.
This means that $\mathcal{Z}_k = \mathcal{Z}_{k'}$ for all $k, k' \in [K]$. Therefore, Eq.~(\ref{eq:zero_patches_specific_k}) holds for all $k\in[K]$, implying all the logit outputs are not affected by the patch $(i,j)$.
\end{proof}

{\bf Remark \ref{remark:hrc_sigmoid_contrastive}.}
\begin{em}
For binary classification tasks where there is a single output logit and {\it sigmoid} applied to obtain the final predictions, HiResCAMs are intrinsically contrastive.
\end{em}
\begin{proof}
    For $K = 2$, i.e., binary classification tasks, one can obtain class probabilities using two setups: (i) $f \in \mathbb{R}^2$ with softmax $\sigma$ activation, or (ii): $f' \in \mathbb{R}$ with sigmoid $\phi$ activation.
    
    Consider ContrastiveCAMs $\contcam{} \in \mathbb{R}^{2 \times d_1 \times d_2}$ for softmax activation, and the corresponding HiResCAMs $\hirescam{} \in \mathbb{R}^{d_1 \times d_2}$ for sigmoid activation. We will prove that HiResCAMs in the sigmoid setup are equivalent to ContrastiveCAMs in the softmax setup.
    
    For the softmax case, we apply Proposition \ref{prop:contcam-correctness} and obtain class probabilities as:
    \begin{align}
        p_f(y = 1 | \rmX)
        = \tilde{f}_1(\rmX)
        &= \left(\exp\left(-\sum^{d_1,d_2}_{i,j=1} \left(\contcamidx{1}{1}\right)_{ij}\right) +\exp\left(- \sum^{d_1, d_2}_{i,j=1} \left(\contcamidx{1}{2}\right){ij} \right)\right)^{-1}. \\
        \intertext{$\contcamidx{k}{k} = \mathbf{0}_{d_1 \times d_2},~\forall k \in [K]$, hence we have:}
        p_f(y = 1 | \rmX)
        &= \left(1+\exp\left(- \sum^{d_1, d_2}_{i,j=1} \left(\contcamidx{1}{2}\right)_{ij} \right)\right)^{-1}  \\
        &= \left(1+\exp\left(\sum^{d_1, d_2}_{i,j=1} \left(\contcamidx{2}{1}\right)_{ij} \right) \right)^{-1},\label{eq:y_1_softmax}
    \end{align}
    and similarly,
    \begin{align}
        p_f(y = 2 | \rmX)
        = \left(1+\exp\left(- \sum^{d_1, d_2}_{i,j=1} \left(\contcamidx{2}{1}\right)_{ij} \right)\right)^{-1}.\label{eq:y_2_softmax}
    \end{align}
    For the sigmoid case, we apply Proposition \ref{prop:hrc-binary-correctness} and obtain class probabilities as:
    \begin{align}
        p_{f'}(y = 1 | \rmX) = 1 - \phi(f') = \phi(-f') = \left(1+\exp\left(\sum^{d_1, d_2}_{i,j=1} \hirescam{ij} \right)\right)^{-1},\label{eq:y_1_sigmoid}
        \intertext{and}
        p_{f'}(y = 2 | \rmX) = \phi(f') = \left(1+\exp\left(- \sum^{d_1, d_2}_{i,j=1} \hirescam{ij} \right)\right)^{-1}.\label{eq:y_2_sigmoid}
    \end{align}
    Comparing Eq.~(\ref{eq:y_1_softmax}) with Eq.~(\ref{eq:y_1_sigmoid}) and Eq.~(\ref{eq:y_2_softmax}) with Eq.~(\ref{eq:y_2_sigmoid}), we observe that HiResCAMs' connection to the model prediction is identical to that of ContrastiveCAMs, hence HiResCAMs in the binary setting are intrinsically contrastive.
\end{proof}

{\bf Proposition \ref{prop:hrc-binary-correctness}} (Correctness of HiResCAMs in Binary Setting){\bf .}
\begin{em}
Given an input $\rmX$, the prediction $\tilde{f}(\rmX)$ of a binary classifier is determined by the corresponding HiResCAM:
    \begin{equation}
        \tilde{f}(\rmX)
        = \left(1+\exp\left(- \sum^{d_1, d_2}_{i,j=1} \hirescam{ij} \right)\right)^{-1}.
    \end{equation}
    \begin{proof}
        In the binary classification setting, $f(\rmX) \in \mathbb{R}$, and $\tilde{f} = \phi(f)$, where $\phi$ denotes {\it sigmoid} activation.
        Applying Eq.~(\ref{eq:logit-connection}), we have:
        \begin{equation}
            \tilde{f}(\rmX) = \phi(f) = (1 + \exp(-f))^{-1} = \left(1+\exp\left(- \sum^{d_1, d_2}_{i,j=1} \hirescam{ij} \right)\right)^{-1}.
        \end{equation}
    \end{proof}
\end{em}

\section{Context-Regularized Binary Cross-Entropy}
\label{sec:crbce}

In the binary classification setting, the neural network outputs a single logit, $f(\rmX) \in \mathbb{R}$, to which the {\it sigmoid} function, denoted $\phi: \mathbb{R} \rightarrow [0, 1]$, is applied to produce the corresponding probability predictions: $\tilde{f} = \phi(f)$.
The corresponding class labels are denoted $y \in \{0, 1\}$.
Additionally, HiResCAMs have shape $\hirescam{} \in \mathbb{R}^{d_1 \times d_2}$, and their connection to logits in Eq.~(\ref{eq:logit-connection}) extends to class probabilities.
This follows from the bijectivity of {\it sigmoid} activation:

\begin{proposition}[Correctness of HiResCAMs in Binary Setting]\label{prop:hrc-binary-correctness}
Given an input $\rmX$, the prediction $\tilde{f}(\rmX)$ of a binary classifier is determined by the corresponding HiResCAM:
\begin{equation}
    \tilde{f}(\rmX)
    = \left(1+\exp\left(- \sum^{d_1, d_2}_{i,j=1} \hirescam{ij} \right)\right)^{-1}.
\end{equation}
\end{proposition}

We leverage this property to apply context regularization in the binary setting:

\begin{definition}[Context-Regularized Binary Cross-Entropy Loss]
Given a training dataset $\mathcal{D}$ and convolutional network $f$, CR-BCE is defined as:
    \begin{align}
	    &\mathcal{L}_{\mathrm{CRBCE}}(\mathcal{D}) = \mathbb{E}_{(\rmX,y,H)\sim \mathcal{D}}[\ell(\rmX,y, H)], \quad\textrm{with}\nonumber\\
        \ell(\rmX,y, H) &= \log \left( 1 + \exp\left( \sum^{d_1, d_2}_{i,j=1} (-1)^y \hirescam{ij} H_{ij} + \underbrace{(1-H_{ij}) |\hirescam{ij}|}_{\textrm{context regularization}} \right) \right).
	\end{align}
\end{definition}
This definition is motivated by noting that the per-sample binary cross-entropy loss $\ell_{BCE}$ can be equivalently expressed as:
\begin{equation}
    \ell_{BCE}(\rmX, y) = -y \log(\tilde{f}) - (1 - y) \log (1 - \tilde{f}) = \log(1 + \exp((-1)^y f)),
\end{equation}
when sigmoid activation is used to obtain class probabilities from logits.

\section{Additional Experimental Results}
\begin{table}
\caption{{\bf Spawrious Context-Robustness Benchmark.}
CR-CE outperforms the standard cross-entropy training, as well as GALS and ABN, across all difficulty levels and context shift settings of the Spawrious dataset.
}
\label{table:spawrious_all}
\centering
\begin{adjustbox}{max width=\textwidth}
\begin{threeparttable}[t]
\begin{tabular}{ccccccc}
\toprule
  \multicolumn{1}{c}{\multirow{2}[3]{*}{\textbf{Spurious Shift Type}}} &
  \multirow{2}[3]{*}{\textbf{\begin{tabular}[c]{@{}c@{}}CAM\\ Size\end{tabular}}} &
  \multirow{2}[3]{*}{\textbf{Approach}} &
  \multicolumn{3}{c}{\textbf{Task Difficulty}} &
  \multirow{2}[3]{*}{\textbf{Average $(\uparrow)$}} \\
  \cmidrule{4-6}
& & & \multicolumn{1}{c}{Easy $(\uparrow)$} & \multicolumn{1}{c}{Medium $(\uparrow)$} & \multicolumn{1}{c}{Hard $(\uparrow)$} & \\
\midrule
\multirow{13}{*}{One-to-One}   & \multirow{6}{*}{$7 \times 7$} &  CE (Baseline) &  $93.30${\tiny $\pm 0.03$} & $83.17${\tiny $\pm 0.47$} & $86.58${\tiny $\pm 1.88$}  & $87.69$ \\
& & GALS & $93.42${\tiny $\pm 0.18$} & $84.39${\tiny $\pm 2.13$} & $87.11${\tiny $\pm 0.27$} & $88.31$ \\
& & EBL & $93.31${\tiny $\pm 0.69$} & $83.91${\tiny $\pm 1.46$} & $86.44${\tiny $\pm 1.45$} & $87.89$ \\
& & CFA & $90.86${\tiny $\pm 0.54$} & $79.46${\tiny $\pm 0.48$} & $81.77${\tiny $\pm 1.73$} & $84.03$ \\
& & Context Masking & $92.36${\tiny $\pm 0.70$} & $80.97${\tiny $\pm 1.59$} & $83.67${\tiny $\pm 3.71$} & $85.66$ \\
& & \bf CR-CE (Ours) & $\mathbf{95.47}${\tiny $\pm 0.24$} & $\mathbf{90.51}${\tiny $\pm 0.16$} & $\mathbf{91.35}${\tiny $\pm 0.34$} & $\mathbf{92.44}$ \\
\cmidrule{2-7}
& \multirow{7}{*}{$14 \times 14$} & CE (Baseline) & $92.83${\tiny $\pm 0.01$} & $80.36${\tiny $\pm 1.46$} & $87.74${\tiny $\pm 0.37$} & $86.98$ \\
& & ABN & $88.01${\tiny $\pm 1.49$} & $80.47${\tiny $\pm 0.61$} & $80.04${\tiny $\pm 2.13$} & $82.84$ \\
& & GALS & $94.72${\tiny $\pm 1.32$} & $85.74${\tiny $\pm 3.96$} & $90.57${\tiny $\pm 0.65$} & $90.35$ \\
& & EBL & $94.20${\tiny $\pm 0.28$} & $82.20${\tiny $\pm 3.33$} & $87.59${\tiny $\pm 1.54$} & $88.00$ \\
& & CFA & $88.98${\tiny $\pm 0.14$} & $73.38${\tiny $\pm 0.87$} & $75.88${\tiny $\pm 1.12$} & $79.41$ \\
& & Context Masking & $93.07${\tiny $\pm 0.40$} & $81.79${\tiny $\pm 2.29$} & $83.91${\tiny $\pm 1.95$} & $86.26$ \\
& & \bf CR-CE (Ours) & $\mathbf{96.35}${\tiny $\pm 0.15$} & $\mathbf{92.24}${\tiny $\pm 0.30$} & $\mathbf{92.89}${\tiny $\pm 0.20$} & $\mathbf{93.83}$ \\
\midrule
\multirow{13}{*}{Many-to-Many} & \multirow{6}{*}{$7 \times 7$} & CE (Baseline) & $90.86${\tiny $\pm 0.54$} & $79.46${\tiny $\pm 0.48$} & $81.52${\tiny $\pm 1.03$} & $83.95$ \\
& & GALS & $93.47${\tiny $\pm 0.36$} & $84.58${\tiny $\pm 0.88$} & $78.74${\tiny $\pm 0.97$} & $85.60$ \\
& & EBL  & $92.43${\tiny $\pm 1.31$} & $87.23${\tiny $\pm 0.84$} & $81.21${\tiny $\pm 0.74$} & $86.96$ \\
& & CFA  & $89.14${\tiny $\pm 0.89$} & $85.68${\tiny $\pm 0.68$} & $81.77${\tiny $\pm 1.73$} & $85.65$ \\
& & Context Masking & $92.08${\tiny $\pm 1.04$} & $85.03${\tiny $\pm 2.37$} & $78.46${\tiny $\pm 1.12$} & $85.19$ \\
& & {\bf CR-CE (Ours)} & $\mathbf{95.55}${\tiny $\pm 0.16$} & $\mathbf{91.25}${\tiny $\pm 0.18$} & $\mathbf{87.49}${\tiny $\pm 0.14$} & $\mathbf{91.43}$ \\
\cmidrule{2-7}
& \multirow{7}{*}{$14 \times 14$} & CE (Baseline) & $88.98${\tiny $\pm 0.14$} & $73.38${\tiny $\pm 0.87$} & $79.76${\tiny $\pm 1.01$} & $80.71$ \\
& & ABN  & $89.65${\tiny $\pm 0.44$} & $72.81${\tiny $\pm 1.32$} & $63.96${\tiny $\pm 6.54$} & $75.47$ \\
& & GALS & $95.03${\tiny $\pm 1.35$} & $87.63${\tiny $\pm 3.80$} & $84.08${\tiny $\pm 1.87$} & $88.91$ \\
& & EBL  & $93.46${\tiny $\pm 0.73$} & $89.05${\tiny $\pm 0.47$} & $79.36${\tiny $\pm 2.83$} & $87.29$ \\
& & CFA  & $87.22${\tiny $\pm 0.78$} & $85.68${\tiny $\pm 0.68$} & $75.88${\tiny $\pm 1.12$} & $82.93$ \\
& & Context Masking & $90.96${\tiny $\pm 0.24$} & $85.70${\tiny $\pm 2.99$} & $80.80${\tiny $\pm 0.71$} & $85.82$ \\
& & {\bf CR-CE (Ours)} & $\mathbf{96.47}${\tiny $\pm 0.06$} & $\mathbf{92.41}${\tiny $\pm 0.29$} & $\mathbf{88.66}${\tiny $\pm 0.52$} & $\mathbf{92.51}$ \\
\bottomrule
\end{tabular}%
\end{threeparttable}
\end{adjustbox}
\end{table}
\begin{table}[t]
	\caption{{\bf Hard-ImageNet Benchmark.}
    Predictions on models trained with CR-CE are least likely to persist when the object is absent from the image.
    Object-focus increases proportional to the CAM resolution, with some performance cost incurred when the target is present alongside the context.
    }
    \label{table:hardinet-results}
    \centering
    \resizebox{\textwidth}{!}{
\begin{tabular}{ccllllll}
\toprule
\multirow{2}[3]{*}{\textbf{\begin{tabular}[c]{@{}c@{}}CAM \\ Size\end{tabular}}} &
\multirow{2}[3]{*}{\textbf{Approach}} &
  \multicolumn{1}{c}{\multirow{2}[3]{*}{\textbf{\begin{tabular}[c]{@{}c@{}}No Shift\\ Accuracy $(\uparrow)$\end{tabular}}}} &
  \multicolumn{3}{c}{\textbf{Accuracy under Target Removal}} &
  \multicolumn{2}{c}{\textbf{Rel.~Foreground Sensitivity}} \\
  \cmidrule(lr){4-6}
  \cmidrule(lr){7-8}
 &
   &
  \multicolumn{1}{c}{} &
  \multicolumn{1}{c}{Gray Mask $(\downarrow)$} &
  \multicolumn{1}{c}{Gray BBOX $(\downarrow)$} &
  \multicolumn{1}{c}{Tile $(\downarrow)$} &
  \multicolumn{1}{c}{$\rho=0.25~(\uparrow)$} &
  \multicolumn{1}{c}{$\rho=0.5 ~(\uparrow)$} \\
\midrule
\multirow{6}{*}{$7 \times 7$} & CE (Baseline) & $95.82${\tiny $\pm 0.08 $}	&	$82.4${\tiny $\pm 0.13$}	&	$79.38${\tiny $\pm 0.2$}	&	$77.11${\tiny $\pm 0.08$}	&	$-0.21${\tiny $\pm 0.01$}	&	$-0.36${\tiny $\pm 0.01$} \\
& GALS & $95.91${\tiny $\pm 0.15 $}	&	$80.93${\tiny $\pm 0.27$}	&	$76.62${\tiny $\pm 1.21$}	&	$75.69${\tiny $\pm 0.73$}	&	$-0.18${\tiny $\pm 0.01$}	&	$-0.31${\tiny $\pm 0.03$} \\
& EBL & $95.87${\tiny $\pm 0.23 $}	&	$77.64${\tiny $\pm 1.66$}	&	$73.20${\tiny $\pm 2.07$}	&	$70.74${\tiny $\pm 1.67$}	&	$-0.11${\tiny $\pm 0.04$}	&	$-0.19${\tiny $\pm 0.06$} \\
& CFA & $\underline{96.00}${\tiny $\pm 0.13$}	&	$82.13${\tiny $\pm 0.35$}	&	$79.24${\tiny $\pm 0.50$}	&	$76.67${\tiny $\pm 0.35$}	&	$-0.20${\tiny $\pm 0.01$}	&	$-0.35${\tiny $\pm 0.01$} \\
& \bf CR-CE (Ours) & $91.42${\tiny $\pm 0.28 $}	& $\mathbf{50.09}${\tiny $\pm 0.89$}	&	$\mathbf{45.78}${\tiny $\pm 0.41$}	&	$\mathbf{46.98}${\tiny $\pm 0.2$}	&	$\mathbf{+0.10}${\tiny $\pm 0.02 $}&	$\mathbf{+0.16}${\tiny $\pm 0.01$} \\
\midrule
\multirow{7}{*}{$14 \times 14$} & CE (Baseline) & $95.82${\tiny $\pm 0.08 $}	&	$81.2${\tiny $\pm 0.61$}	&	$78.36${\tiny $\pm 0.28$}	&	$77.64${\tiny $\pm 0.2$}	&	$-0.22${\tiny $\pm 0.01$}	&	$-0.34${\tiny $\pm 0.003$} \\
& ABN & $95.24${\tiny $\pm 0.54 $}	&	$80.84${\tiny $\pm 0.15$}	&	$73.47${\tiny $\pm 1.62$}	&	$73.33${\tiny $\pm 0.23$}	&	$-0.34${\tiny $\pm 0.02$}	&	$-0.44${\tiny $\pm 0.01$} \\
& GALS & $95.82${\tiny $\pm 0.15 $}	&	$78.00${\tiny $\pm 3.54$}	&	$74.27${\tiny $\pm 4.35$}	&	$73.82${\tiny $\pm 3.78$}	&	$-0.12${\tiny $\pm 0.08$}	&	$-0.23${\tiny $\pm 0.14$} \\
& EBL & $95.24${\tiny $\pm 0.15$}	&	$70.31${\tiny $\pm 4.46$}	&	$67.16${\tiny $\pm 3.45$}	&	$65.91${\tiny $\pm 5.33$}	&	$-0.04${\tiny $\pm 0.03$}	&	$-0.05${\tiny $\pm 0.14$} \\
& CFA & $\underline{96.00}${\tiny $\pm 0.13 $}	&	$81.42${\tiny $\pm 0.50$}	&	$78.22${\tiny $\pm 0.56$}	&	$76.76${\tiny $\pm 0.73$}	&	$-0.18${\tiny $\pm 0.02$}	&	$-0.36${\tiny $\pm 0.06$} \\
& \bf CR-CE (Ours) & $88.53${\tiny $\pm 0.00 $}	&	$\mathbf{37.07}${\tiny $\pm 0.48$}	&	$\mathbf{32.44}${\tiny $\pm 0.15$}	&	$\mathbf{32.76}${\tiny $\pm 0.78$}	&	$\mathbf{+0.17}${\tiny $\pm 0.01$}	&	$\mathbf{+0.23}${\tiny $\pm 0.01$} \\
\bottomrule
\end{tabular}
}
\end{table}

\subsection{Spawrious Benchmark on Various Context-Shift Configurations}
\label{sec:spawrious_all_configuration_results}
The Spawrious dataset comprises of four dog breeds spread across six contexts, with three difficulty levels based on the object-context combinations available during training, and two different configurations of context shift: (i) one-to-one shifts, where each class is correlated with one context, and (ii) many-to-many shifts, where groups of classes are correlated with groups of contexts, the latter being observed broadly in practice \citep{lynch2023spawrious}.
We include results on both one-to-one and many-to-many shifts in Table \ref{table:spawrious_all}.
Consistent with our findings in Section \ref{sec:crce-improves-robustness}, we observe that CR-CE trained models outperform the baseline and existing explanation-guided approaches in context-shifted accuracy across all difficulty levels.

\subsection{Evaluating Context Dependence using Hard-ImageNet}
\label{sec:hardinet_evals}

The Hard-ImageNet benchmark \citep{moayeri2022hard} assesses model reliance on context by comparing performance on images with and without the target object.
Ideally, the model should predict the true class in the presence of the object, and a random class in the absence of the object.
In other words, the prediction should not be based on the context alone.
The Hard-ImageNet suite also reports Relative Foreground Sensitivity (RFS) \citep{moayeri2022comprehensive}, which compares accuracy reduction when gaussian noise (with magnitude denoted $\rho$) is added to either the object or the context region; higher RFS signifies greater object-dependence.

\begin{figure}[t]
    \centering
    \includegraphics[width=\textwidth]{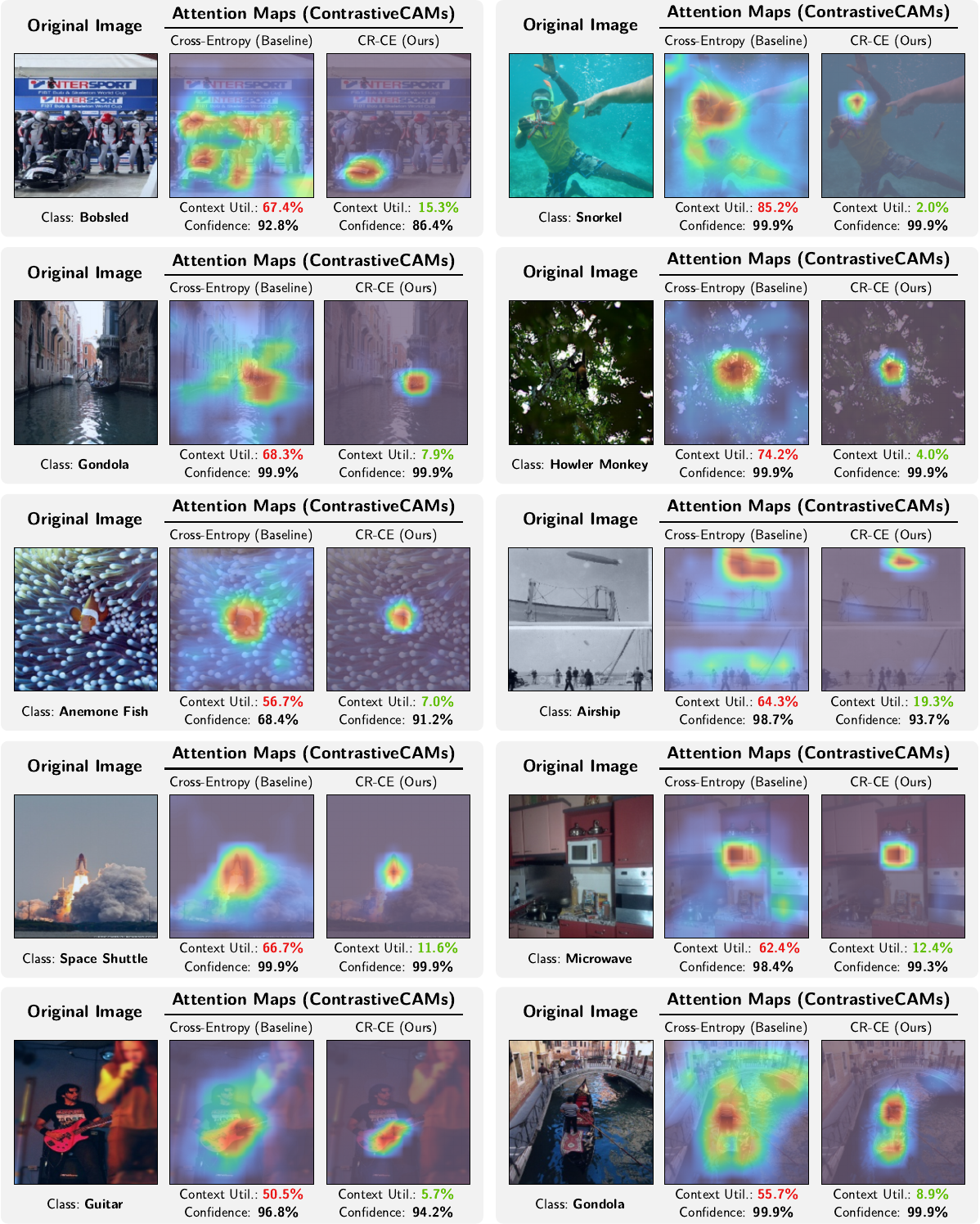}
    \caption{{\bf ImageNet-BG} and {\bf Hard-ImageNet.}
    Extended qualitative evaluations of object-context dependence by cross-entropy and CR-CE trained models.
    Images belonging to the {\it Snorkel} and {\it Howler Monkey} classes correspond to the Hard-ImageNet validation set, whereas the rest correspond to the ImageNet-BG test set.
    Across each of the examples, we observe the context dependence in models trained using cross-entropy loss is significantly mitigated by training with CR-CE.}
    \label{fig:imagenet_hard_bg_examples}
\end{figure}

Evaluation of CR-CE on the Hard-ImageNet and comparison with other methods are reported in Table \ref{table:hardinet-results}.
We observe that the CR-CE framework meaningfully outperforms each of the compared methods on key robustness metrics assessed by the Hard-ImageNet benchmark. Example images of Hard-ImageNet demonstrating the effectiveness of CR-CE on suppressing context dependence can be find in Figure \ref{fig:crce-imagenetbg} and Figure \ref{fig:imagenet_hard_bg_examples}, see the classes {\it Baseball Player}, {\it Snorkel}, and {\it Howler Monkey}.

\begin{remark}
Hard-ImageNet includes a lot of context-heavy images where object regions comprise only $13.96\%$ of the image on average, leading to a bigger reduction of accuracy in the accuracy-robustness tradeoff. 
\end{remark}

\subsection{Comparison with Explanation-Guided Learning (EGL) Methods}
\label{sec:egl-comparison}

\begin{table}[t]
\caption{{\bf Context-Contribution Ratios.}
We report the context contribution ratios across each of the assessed approaches.
CR-CE trained models place a greater emphasis on object-specific regions to determine predictions and learn to ignore cues from context regions through the process of training.
}
\label{table:contribution-ratio-appendix}
\centering

\begin{adjustbox}{max width=\textwidth}
\begin{threeparttable}
\begin{tabular}{cccccc}
\toprule
\multirow{2}[3]{*}{\textbf{\begin{tabular}[c]{@{}c@{}}CAM \\ Size\end{tabular}}} &
\multirow{2}[3]{*}{\textbf{Approach}} & 
\multicolumn{4}{c}{\textbf{Context Contribution Ratio $R\%$ $(\downarrow)$}}                                                    \\
\cmidrule{3-6}
      &                           &  Waterbirds-100  & Spawrious\tnote{\textdaggerdbl} &  ImageNet-BG & Hard-ImageNet  \\
      \midrule
      \multirow{6}{*}{$7 \times 7$} & CE (Baseline) & $46.08${\tiny $\pm 0.43$} &	$44.79${\tiny $\pm 0.3$}	& $31.83$\tnote{\textdagger}	& $68.53${\tiny $\pm 0.04$} \\
      & GALS & $36.2${\tiny $\pm 0.92$}  & $28.06${\tiny $\pm 1.39$}	& $29.75${\tiny $\pm 0.5$}	& $63.64${\tiny $\pm 2.03$} \\
      & EBL & $56.02${\tiny $\pm 5.17$} &	$21.37${\tiny $\pm 1.45$}	& $25.35${\tiny $\pm 2.06$}	& $57.24${\tiny $\pm 3.91$} \\
      & CFA & $17.06${\tiny $\pm 0.16$} &	$43.1${\tiny $\pm 2.69$}	& $32.49${\tiny $\pm 0.07$}	& $57.22${\tiny $\pm 0.50$} \\
      & Context Masking & $22.48${\tiny $\pm 1.72$} &	$48.11${\tiny $\pm 2.22$}	& $32.86${\tiny $\pm 0.02$}	& $70.63${\tiny $\pm 0.06$} \\
      & \bf CR-CE (Ours) & $\mathbf{12.86}${\tiny $\pm 0.09$} &	$\mathbf{14.89}${\tiny $\pm 0.2$}	& $\mathbf{14.85}${\tiny $\pm 0.00$}	& $\mathbf{35.85}${\tiny $\pm 0.26$} \\
      \midrule
      \multirow{7}{*}{$14 \times 14$}  & CE (Baseline) & $58.75${\tiny $\pm 0.12$} &	$47.32${\tiny $\pm 0.09$}	& $28.27${\tiny $\pm 0.27$}	& $66.5${\tiny $\pm 0.07$} \\
      & ABN & $34.92${\tiny $\pm 19.38$} &$49.89${\tiny $\pm 16.99$}	& $33.37${\tiny $\pm 1.01$}	& $69.06${\tiny $\pm 4.02$} \\
      & GALS & $35.7${\tiny $\pm 2.42$}  & $17.77${\tiny $\pm 2.07$}	& $24.14${\tiny $\pm 0.13$}	& $56.87${\tiny $\pm 9.37$} \\
      & EBL & $48.21${\tiny $\pm 2.44$} &	$29.82${\tiny $\pm 1.06$}	& $22.94${\tiny $\pm 0.83$}	& $48.63${\tiny $\pm 9.68$} \\
      & CFA & $16.08${\tiny $\pm 0.88$} &	$46.66${\tiny $\pm 0.58$}	& $28.42${\tiny $\pm 0.02$}	& $47.89${\tiny $\pm 1.56$} \\
      & Context Masking & $25.24${\tiny $\pm 1.27$} &	$38.44${\tiny $\pm 1.91$}	& $27.64${\tiny $\pm 0.03$}	& $69.35${\tiny $\pm 0.2$} \\
      & \bf CR-CE (Ours) & $\mathbf{9.54}${\tiny $\pm 0.28$}  & $\mathbf{10.98}${\tiny $\pm 0.12$}	& $\mathbf{9.32}${\tiny $\pm 0.06$}	& $\mathbf{31.12}${\tiny $\pm 0.1$} \\
      \bottomrule
\end{tabular}

\begin{tablenotes}
    \small
    \item [\textdagger] No standard deviation is reported; we use pre-trained weights from \texttt{torchvision} without modification.
    \item [\textdaggerdbl] We report metrics on the hardest subtask and difficulty (M2M-hard) within the Spawrious benchmark.
\end{tablenotes}
\end{threeparttable}
\end{adjustbox}
\end{table}

\begin{figure}[h!]
    \centering
    \includegraphics[width=\textwidth]{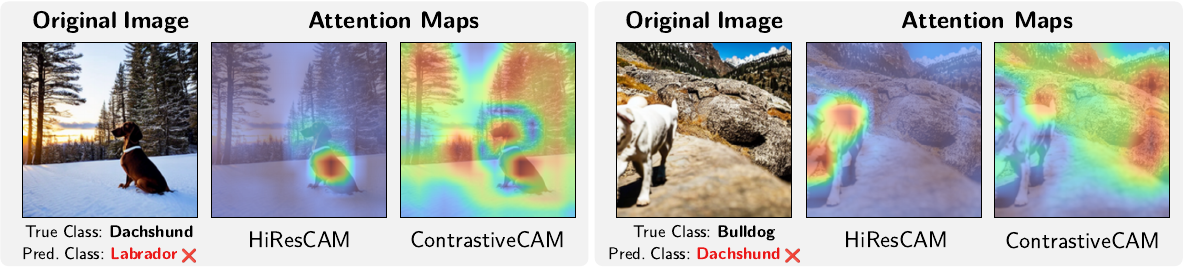}
    \caption{{\bf Superficially Aligned Attribution Maps.}
    On a ResNet-50 trained with the GALS framework, we observe that HiResCAMs suggest a focus on the object regions; however, the model predicts the context-correlated class observed during training.
    ContrastiveCAMs faithfully reveal that the context contribution is in fact the primary driver of the prediction.
    This highlights the essential role that explanation faithfulness plays in ensuring improvements to context robustness in the explanation-guided learning paradigm.
    }
    \label{fig:gals_cam_comparison}
\end{figure}

Typically, explanation-guided learning frameworks \citep{gao2024going} construct a multi-task optimization objective with a per-sample loss of the following form:
\begin{equation}\label{eq:egl-framework}
    \ell_{\mathrm{EGL}} = \ell_{\mathrm{CE}} + \lambda \cdot \ell_{\mathrm{EG}},
    \quad
    \textrm{with gradient}
    \quad
    \nabla \ell_{\mathrm{EGL}} = \nabla \ell_{\mathrm{CE}} + \lambda \nabla \ell_{\mathrm{EG}}.
\end{equation}
Here, $\ell_{\mathrm{EG}}$ is a secondary explanation-guided goal designed to suppress context contributions.
The specific form of this term varies between approaches \citep{sarvamangala2022convolutional, rao2023studying, ismail2021improving, petryk2022gals, fukui2019attention}.

For a given $(i,j)$-patch corresponding to context, gradients of the form in Eq.~(\ref{eq:egl-framework}) create a tug-of-war dynamic: the first term seeks to minimize $\ell_{\mathrm{CE}}$ at the cost of $\ell_{\mathrm{EG}}$, and vice versa.
To manage this, the additional hyperparameter $\lambda$ is introduced and must be carefully tuned to prevent poor classification accuracy.
In contrast, CR-CE replaces the mechanism in cross-entropy that increases contributions from context regions, preventing competing gradients (taken w.r.t.~the specific patch), hence resulting in a more pronounced object-focus effect.

The advantage of this is evident in the Waterbirds-100 dataset.
Since it is a binary classification task, the GradCAMs used by GALS coincide with ContrastiveCAMs used by CR-CE.
The only difference between the two approaches lies in how the feature attribution method is incorporated in the model training procedure.
In this direct comparison, CR-CE continues to outperform GALS, see Table \ref{table:results_overview}.

\paragraph{Explanation Correctness is Key to Robustness Improvements.}

In Table \ref{table:results_overview}, we observe that some prior explanation-guided methods may even have an inferior performance than the CE training  baselines, especially when the tasks become complicated.
To investigate this performance drift, we measure the context-contribution ratios for those approaches in Table \ref{table:contribution-ratio-appendix}.
We observe that models trained by those  explanation-guided methods often have greater context dependence, indicating the ineffectiveness of these methods in guiding the feature attributions towards object regions. We hypothesize that this behavior is largely due to their reliance on classical feature attribution method (e.g. Grad-CAM and HiResCAM), which
suffer from the unfaithfulness issue theoretically described in Theorem \ref{thm:hirescam}. Figure \ref{fig:gals_cam_comparison} shows typical examples where prior explanation-guided methods may be successful in guiding those classical CAMs to focus on the object regions but fail in doing so for the more faithful ContrastiveCAMs.

\subsection{CR-CE is Tolerant to Imperfect Object-Context Segmentation}
\label{sec:crce-tolerance-noisy-masks}
In our experiments with Spawrious and ImageNet datasets in Table \ref{table:results_overview}, we already used AI-predicted object-context segmentation, obtained from SAM3 \citep{carion2026sam} and MAEs \citep{he2022masked}, which are approximations and are inevitably imperfect.
Here, 
we further assess CR-CE's tolerance to noise by adding additional noise to these object-context masks.
Specifically, we perform random affine transformations (up to $5^\circ$ of rotation, $\pm 7.5\%$ translation, and $\pm 30\%$ scaling) on each of those masks $H$.
This augmentation results in $\approx 50\%$ mIoU of the original object-context masks to their corrupted counterparts.
We train CR-CE using  these corrupted masks, and report our results in Table \ref{table:noisy_masks}.
We can see that, even with these noisy  object-context segmentation, CR-CE can still predict well with minor degradation in the performance. 

\begin{table}[t]
\caption{
{\bf Context-Shift Robustness using Noisy Masks (NM).}
Despite the additional noise introduced to object-context masks, CR-CE remains effective at improving context-shift robustness.
CR-CE retains its utility in scenarios where accurate masks are challenging or expensive to obtain.
}
\label{table:noisy_masks}
\centering
\begin{adjustbox}{max width=\textwidth}
\begin{threeparttable}[h]
\begin{tabular}{ccccccccc}
\toprule
\multirow{2}[3]{*}{\textbf{\begin{tabular}[c]{@{}c@{}}CAM \\ Size\end{tabular}}} &
\multirow{2}[3]{*}{\bf Approach} &
  \multicolumn{2}{c}{\bf Waterbirds-100} &
  \multicolumn{2}{c}{\bf ImageNet/ImageNet-BG} &
  \multicolumn{3}{c}{\bf Spawrious} \\
  \cmidrule(lr){3-4}
  \cmidrule(lr){5-6}
  \cmidrule(lr){7-9}
 &
  \multicolumn{1}{c}{} &
  \multicolumn{1}{c}{No Shift $(\uparrow)$} &
  \multicolumn{1}{c}{Shifted $(\uparrow)$} &
  \multicolumn{1}{c}{No Shift $(\uparrow)$} &
  \multicolumn{1}{c}{Shifted $(\uparrow)$} &
  \multicolumn{1}{c}{Easy $(\uparrow)$} &
  \multicolumn{1}{c}{Medium $(\uparrow)$} &
  \multicolumn{1}{c}{Hard $(\uparrow)$} \\
\midrule
\multirow{3}{*}{$7 \times 7$} & CE (Baseline) & $\underline{98.66}${\tiny $\pm 0.10$} & $41.74${\tiny $\pm 4.15$} & $\underline{82.92}$\tnote{\textdagger} & $74.83$\tnote{\textdagger} & $90.86${\tiny $\pm 0.54$} & $79.46${\tiny $\pm 0.48$} & $81.52${\tiny $\pm 1.03$} \\
& {\bf CR-CE (Ours) + NM} & $92.10${\tiny $\pm 0.22$} & $72.61${\tiny $\pm 1.26$} & $81.14${\tiny $\pm 0.02$} & $74.89${\tiny $\pm 0.07$} & $93.07${\tiny $\pm 0.32$} & $87.75${\tiny $\pm 0.42$} & $82.19${\tiny $\pm 0.16$} \\
& {\bf CR-CE (Ours)} & $94.38${\tiny $\pm 1.21$} & $\mathbf{73.30}${\tiny $\pm 4.43$} & $81.85${\tiny $\pm 0.03$} & $\mathbf{75.75}${\tiny $\pm 0.08$} & $\mathbf{95.55}${\tiny $\pm 0.16$} & $\mathbf{91.25}${\tiny $\pm 0.18$} & $\mathbf{87.49}${\tiny $\pm 0.14$} \\
\midrule
\multirow{3}{*}{$14 \times 14$} & CE (Baseline) & $\underline{98.48}${\tiny $\pm 0.09$} & $38.40${\tiny $\pm 1.79$} & $82.33${\tiny $\pm 0.10$} & $74.72${\tiny $\pm 0.06$} & $88.98${\tiny $\pm 0.14$} & $73.38${\tiny $\pm 0.87$} & $79.76${\tiny $\pm 1.01$} \\
& {\bf CR-CE (Ours) + NM} & $90.67${\tiny $\pm 2.42$} & $76.12${\tiny $\pm 3.24$} & $79.83${\tiny $\pm 0.06$} & $75.24${\tiny $\pm 0.02$} & $93.07${\tiny $\pm 0.51$} & $85.64${\tiny $\pm 0.17$} & $80.59${\tiny $\pm 0.84$} \\
& {\bf CR-CE (Ours)} & $94.36${\tiny $\pm 0.12$} & $\mathbf{79.13}${\tiny $\pm 0.77$} & $81.25${\tiny $\pm 0.03$} & $\mathbf{76.84}${\tiny $\pm 0.12$} & $\mathbf{96.47}${\tiny $\pm 0.06$} & $\mathbf{92.41}${\tiny $\pm 0.29$} & $\mathbf{88.66}${\tiny $\pm 0.52$} \\
\bottomrule
\end{tabular}
\begin{tablenotes}
\small
    \item [\textdagger] No std.~dev.~is reported; we use pre-trained weights from \texttt{torchvision} without modification.
\end{tablenotes}

\end{threeparttable}
\end{adjustbox}
\end{table}

\subsection{CR-CE is Effective Across Architectures and Scales}
\label{sec:crce-effective-across-archs}
Here we provide additional experimental validations of the CR-CE training framework on different network architectures, including
 EfficientNetV2 \citep{tan2019efficientnet} and ResNet-18.
The results are reported in Table \ref{table:efficientnet_v2} and \ref{table:resnet18} respectively, demonstrate the effectiveness of CR-CE across variations in architectures and scales.

\begin{table}[t]
\caption{
{\bf Context-Shift Robustness On EfficientNetV2.}
CR-CE improves context-shift robustness on varied convolutional backbones.
Following EfficientNetV2's \citep{tan2019efficientnet} training recipe, we adopt a larger batch size of $4096$ and increase the learning rate to $10^{-2}$ for ImageNet training.
}
\label{table:efficientnet_v2}
\centering
\begin{adjustbox}{max width=\textwidth}
\begin{threeparttable}[h]
\begin{tabular}{ccccccccc}
\toprule
\multirow{2}[3]{*}{\textbf{\begin{tabular}[c]{@{}c@{}}CAM \\ Size\end{tabular}}} &
\multirow{2}[3]{*}{\bf Approach} &
  \multicolumn{2}{c}{\bf Waterbirds-100} &
  \multicolumn{2}{c}{\bf ImageNet/ImageNet-BG} &
  \multicolumn{3}{c}{\bf Spawrious} \\
  \cmidrule(lr){3-4}
  \cmidrule(lr){5-6}
  \cmidrule(lr){7-9}
 &
  \multicolumn{1}{c}{} &
  \multicolumn{1}{c}{No Shift $(\uparrow)$} &
  \multicolumn{1}{c}{Shifted $(\uparrow)$} &
  \multicolumn{1}{c}{No Shift $(\uparrow)$} &
  \multicolumn{1}{c}{Shifted $(\uparrow)$} &
  \multicolumn{1}{c}{Easy $(\uparrow)$} &
  \multicolumn{1}{c}{Medium $(\uparrow)$} &
  \multicolumn{1}{c}{Hard $(\uparrow)$} \\
\midrule
\multirow{2}{*}{$7 \times 7$} & CE (Baseline) & $91.08${\tiny $\pm 0.05$} & $39.31${\tiny $\pm 3.01$} & $\underline{83.71}$\tnote{\textdagger} & $73.58$\tnote{\textdagger} & $92.62${\tiny $\pm 0.41$} & $81.49${\tiny $\pm 0.29$} & $77.50${\tiny $\pm 0.91$} \\
& {\bf CR-CE (Ours)} & $\underline{94.31}${\tiny $\pm 1.93$} & $\mathbf{78.55}${\tiny $\pm 5.01$} & $83.36${\tiny $\pm 0.16$} & $\mathbf{74.46}${\tiny $\pm 0.08$} & $\mathbf{93.86}${\tiny $\pm 0.62$} & $\mathbf{84.23}${\tiny $\pm 0.67$} & $\mathbf{88.29}${\tiny $\pm 0.22$} \\
\midrule
\multirow{2}{*}{$14 \times 14$} & CE (Baseline) & $84.64${\tiny $\pm 0.14$} & $38.88${\tiny $\pm 1.31$} & $\underline{84.00}${\tiny $\pm 0.07$} & $74.53${\tiny $\pm 0.07$} & $93.11${\tiny $\pm 0.42$} & $80.27${\tiny $\pm 2.06$} & $79.30${\tiny $\pm 1.17$} \\
& {\bf CR-CE (Ours)} & $\underline{94.02}${\tiny $\pm 1.15$} & $\mathbf{80.43}${\tiny $\pm 5.01$} & $83.67${\tiny $\pm 0.03$} & $\mathbf{75.20}${\tiny $\pm 0.04$} & $\mathbf{94.84}${\tiny $\pm 0.33$} & $\mathbf{86.98}${\tiny $\pm 0.04$} & $\mathbf{88.85}${\tiny $\pm 0.55$} \\
\bottomrule
\end{tabular}
\begin{tablenotes}
\small
    \item [\textdagger] No std.~dev.~is reported; we use pre-trained weights from \texttt{torchvision} without modification.
\end{tablenotes}
\end{threeparttable}
\end{adjustbox}
\end{table}

\begin{table}[t]
\caption{
{\bf Context-Shift Robustness On ResNet-18.}
CR-CE improves context-shift robustness on the smaller ResNet-18 backbone.
We observe a more pronounced trade-off between in-distribution and context-shifted accuracy due to the smaller size of the backbone.
In contrast, this trade-off is minimized by the larger capacity of the EfficientNetV2 backbone, see Table \ref{table:efficientnet_v2}.
}
\label{table:resnet18}
\centering
\begin{adjustbox}{max width=\textwidth}
\begin{threeparttable}[h]
\begin{tabular}{ccccccccc}
\toprule
\multirow{2}[3]{*}{\textbf{\begin{tabular}[c]{@{}c@{}}CAM \\ Size\end{tabular}}} &
\multirow{2}[3]{*}{\bf Approach} &
  \multicolumn{2}{c}{\bf Waterbirds-100} &
  \multicolumn{2}{c}{\bf ImageNet/ImageNet-BG} &
  \multicolumn{3}{c}{\bf Spawrious} \\
  \cmidrule(lr){3-4}
  \cmidrule(lr){5-6}
  \cmidrule(lr){7-9}
 &
  \multicolumn{1}{c}{} &
  \multicolumn{1}{c}{No Shift $(\uparrow)$} &
  \multicolumn{1}{c}{Shifted $(\uparrow)$} &
  \multicolumn{1}{c}{No Shift $(\uparrow)$} &
  \multicolumn{1}{c}{Shifted $(\uparrow)$} &
  \multicolumn{1}{c}{Easy $(\uparrow)$} &
  \multicolumn{1}{c}{Medium $(\uparrow)$} &
  \multicolumn{1}{c}{Hard $(\uparrow)$} \\
\midrule
\multirow{2}{*}{$7 \times 7$} & CE (Baseline) & $79.96${\tiny $\pm 12.03$} & $38.00${\tiny $\pm 3.79$} & $\underline{73.49}$\tnote{\textdagger} & $58.38$\tnote{\textdagger} & $80.89${\tiny $\pm 0.92$} & $59.34${\tiny $\pm 0.73$} & $52.16${\tiny $\pm 1.27$} \\
& {\bf CR-CE (Ours)} & $\underline{88.48}${\tiny $\pm 0.09$} & $\mathbf{53.55}${\tiny $\pm 10.44$} & $72.40${\tiny $\pm 0.05$} & $\mathbf{62.61}${\tiny $\pm 0.02$} & $\mathbf{90.77}${\tiny $\pm 0.07$} & $\mathbf{75.93}${\tiny $\pm 0.56$} & $\mathbf{68.99}${\tiny $\pm 0.29$} \\
\midrule
\multirow{2}{*}{$14 \times 14$} & CE (Baseline) & $\underline{97.59}${\tiny $\pm 0.36$} & $30.41${\tiny $\pm 8.90$} & $\underline{75.02}${\tiny $\pm 0.10$} & $60.43${\tiny $\pm 0.06$} & $74.84${\tiny $\pm 2.08$} & $46.14${\tiny $\pm 2.45$} & $37.81${\tiny $\pm 1.14$} \\
& {\bf CR-CE (Ours)} & $88.25${\tiny $\pm 0.22$} & $\mathbf{59.95}${\tiny $\pm 7.97$} & $71.44${\tiny $\pm 0.06$} & $\mathbf{63.37}${\tiny $\pm 0.04$} & $\mathbf{91.64}${\tiny $\pm 0.26$} & $\mathbf{79.42}${\tiny $\pm 0.23$} & $\mathbf{70.68}${\tiny $\pm 1.44$} \\
\bottomrule
\end{tabular}
\begin{tablenotes}
\small
    \item [\textdagger] No std.~dev.~is reported; we use pre-trained weights from \texttt{torchvision} without modification.
\end{tablenotes}
\end{threeparttable}
\end{adjustbox}
\end{table}

\section{About Data Augmentation with Object-Context Mixing}
\label{sec:crce-comparison-with-ocm}
A straightforward yet often effective approach to improving robustness to context shifts is to augment the training dataset with synthetic images that combine objects with diverse contexts \citep{gao2023out}. This approach is orthogonal to the methods considered in this paper, including explanation-guided methods, as it leaves the training procedure unchanged and simply introduces additional augmented training samples. Therefore, it can be readily combined with our proposed method. However, its effectiveness may be limited by the finite set of object–context combinations observed during training, potentially leaving the model vulnerable to unseen context shifts at evaluation time, particularly adversarial ones. In contrast, our proposed training framework suppresses the influence of context regions, not limited by the artificial object–context combinations seen during training, potentially providing greater robustness to out-of-distribution scenarios.

Nevertheless, we empirically evaluate this approach by augmenting each training image with an additional image in which the object is paired with a context uniformly sampled from the non-target classes, thereby doubling the size of the training dataset.
We report the effect of pairing this augmentation with both CE and CR-CE in Table \ref{table:object_context_mixing}. We can see that the data augmentation technique performs well in simple tasks, such as Waterbirds, but becomes worse when the task becomes harder.

\begin{table}[t]
\caption{
{\bf Context-Shift Robustness using Object-Context Augmentation.}
Generating object-context combinations using copy-paste augmentation only improves robustness in the Waterbirds setting, when object-context augmentation simulates the test set.
In larger and more realistic context-shift scenarios, augmenting contexts does not produce enough signal to suppress context contributions.
}
\label{table:object_context_mixing}
\centering
\begin{adjustbox}{max width=\textwidth}
\begin{threeparttable}[h]
\begin{tabular}{ccccccccc}
\toprule
\multirow{2}[3]{*}{\textbf{\begin{tabular}[c]{@{}c@{}}CAM \\ Size\end{tabular}}} &
\multirow{2}[3]{*}{\bf Approach} &
  \multicolumn{2}{c}{\bf Waterbirds-100} &
  \multicolumn{2}{c}{\bf ImageNet/ImageNet-BG} &
  \multicolumn{3}{c}{\bf Spawrious} \\
  \cmidrule(lr){3-4}
  \cmidrule(lr){5-6}
  \cmidrule(lr){7-9}
 &
  \multicolumn{1}{c}{} &
  \multicolumn{1}{c}{No Shift $(\uparrow)$} &
  \multicolumn{1}{c}{Shifted $(\uparrow)$} &
  \multicolumn{1}{c}{No Shift $(\uparrow)$} &
  \multicolumn{1}{c}{Shifted $(\uparrow)$} &
  \multicolumn{1}{c}{Easy $(\uparrow)$} &
  \multicolumn{1}{c}{Medium $(\uparrow)$} &
  \multicolumn{1}{c}{Hard $(\uparrow)$} \\
\midrule
\multirow{4}{*}{$7 \times 7$} & CE (Baseline) & $\underline{98.66}${\tiny $\pm 0.10$} & $41.74${\tiny $\pm 4.15$} & $\underline{82.92}$\tnote{\textdagger} & $74.83$\tnote{\textdagger} & $90.86${\tiny $\pm 0.54$} & $79.46${\tiny $\pm 0.48$} & $81.52${\tiny $\pm 1.03$} \\
& CE + OCM & $96.01${\tiny $\pm 0.11$} & $90.64${\tiny $\pm 0.81$} & $80.38${\tiny $\pm 0.13$} & $72.80${\tiny $\pm 0.04$} & $79.65${\tiny $\pm 0.39$} & $49.15${\tiny $\pm 2.01$} & $42.99${\tiny $\pm 1.01$} \\
& {\bf CR-CE (Ours)} & $94.38${\tiny $\pm 1.21$} & $73.30${\tiny $\pm 4.43$} & $81.85${\tiny $\pm 0.03$} & $\mathbf{75.75}${\tiny $\pm 0.08$} & $\mathbf{95.55}${\tiny $\pm 0.16$} & $\mathbf{91.25}${\tiny $\pm 0.18$} & $\mathbf{87.49}${\tiny $\pm 0.14$} \\
& {\bf CR-CE (Ours) + OCM} & $95.35${\tiny $\pm 1.02$} & $\mathbf{92.63}${\tiny $\pm 0.97$} & $79.52${\tiny $\pm 0.03$} & $73.79${\tiny $\pm 0.01$} & $94.40${\tiny $\pm 0.09$} & $87.95${\tiny $\pm 0.61$} & $76.96${\tiny $\pm 0.60$} \\
\midrule
\multirow{4}{*}{$14 \times 14$} & CE (Baseline) & $\underline{98.48}${\tiny $\pm 0.09$} & $38.40${\tiny $\pm 1.79$} & $82.33${\tiny $\pm 0.10$} & $74.72${\tiny $\pm 0.06$} & $88.98${\tiny $\pm 0.14$} & $73.38${\tiny $\pm 0.87$} & $79.76${\tiny $\pm 1.01$} \\
& CE + OCM & $95.51${\tiny $\pm 0.22$} & $92.73${\tiny $\pm 1.06$} & $81.08${\tiny $\pm 0.14$} & $73.70${\tiny $\pm 0.05$} & $70.69${\tiny $\pm 2.47$} & $40.84${\tiny $\pm 0.74$} & $32.17${\tiny $\pm 5.06$} \\
& {\bf CR-CE (Ours)} & $94.36${\tiny $\pm 0.12$} & $79.13${\tiny $\pm 0.77$} & $81.25${\tiny $\pm 0.03$} & $\mathbf{76.84}${\tiny $\pm 0.12$} & $\mathbf{96.47}${\tiny $\pm 0.06$} & $\mathbf{92.41}${\tiny $\pm 0.29$} & $\mathbf{88.66}${\tiny $\pm 0.52$} \\
& {\bf CR-CE (Ours) + OCM} & $94.88${\tiny $\pm 0.29$} & $\mathbf{93.36}${\tiny $\pm 0.88$} & $78.85${\tiny $\pm 0.07$} & $74.75${\tiny $\pm 0.03$} & $95.30${\tiny $\pm 0.24$} & $88.55${\tiny $\pm 0.36$} & $77.51${\tiny $\pm 0.41$} \\
\bottomrule
\end{tabular}
\begin{tablenotes}
\small
    \item [\textdagger] No std.~dev.~is reported; we use pre-trained weights from \texttt{torchvision} without modification.
\end{tablenotes}
\end{threeparttable}
\end{adjustbox}
\end{table}

\section{Training Details}
\label{sec:training-detail}

\paragraph{Hyperparameter Settings.}
For all methods, we use Momentum SGD \citep{polyak1964some} with a momentum parameter $0.9$. The hyperparameter settings shown in Table \ref{table:hyperparameters} are applied  across all methods.
GALS \citep{petryk2022gals}, ABN \citep{fukui2019attention} and Energy loss \citep{rao2023studying}
each have an additional hyperparameter $\lambda$ which weights the regularization strength of spatial feature attributions, which we keep at their default value of 1.

\begin{table}[h!]
\centering
\caption{
\textbf{Hyperparameters} are dataset-specific and kept consistent across approaches.
Hard-ImageNet and ImageNet-1k are trained for fewer epochs since we fine-tune ImageNet-trained models.
We increase the batch size as number of classes (2, 4, 15, and 1000 respectively) increases.}
\label{table:hyperparameters}
\resizebox{\textwidth}{!}{%
\begin{tabular}{ccccccc}
\toprule
\textbf{Dataset} & \textbf{Epochs} & \textbf{Batch Size} & \textbf{LR} & \textbf{Momentum} & \textbf{Weight Decay} & \textbf{Scheduler} \\
\midrule
Waterbirds-100            & $150$ & $64$  & $10^{-3}$ & \multirow{4}{*}{$0.9$} & $10^{-4}$ & -                \\
Spawrious                 & $150$ & $128$ & $10^{-3}$ &                        & $10^{-4}$ & -                \\
Hard-ImageNet             & $100$ & $256$ & $5 \cdot 10^{-3}$ &                & $10^{-4}$ & -                \\
ImageNet-1k / ImageNet-BG & $20$  & $512$ & $5 \cdot 10^{-4}$ &                & $10^{-5}$ & Cosine Annealing \\
\bottomrule
\end{tabular}%
}
\end{table}

\paragraph{Computational Resources.}
We train each of our models using a single NVIDIA H100 GPU.
The most efficient loss function, cross-entropy, took a maximum of $5.5$ hours to train ResNet-50 on ImageNet, and a minimum of $20$ minutes on the Waterbirds dataset.
Training with CR-CE took a maximum of $6.1$ hours per ResNet-50 run on the ImageNet dataset, and a minimum of $20$ minutes on the Waterbirds dataset.
Each of the other methods approximately took between $8$ and $12$ hours to train a ResNet-50 on ImageNet.
Each of the methods used a maximum of $65$ GB of VRAM during training on the ImageNet dataset, with gradient accumulation applied as necessary to stay within the VRAM limit of the GPU.
We trained a total of $582$ models: $3$ independent runs for each approach and backbone on $9$ different training sets ($6$ for Spawrious, $1$ each for Waterbirds, ImageNet-BG, and Hard-ImageNet), with $2$ CAM resolutions for all but ABNs.
ABN increases the CAM resolution as part of its architecture design; we do not reduce it to $7 \times 7$.

\paragraph{Reproducibility.} 
The source code 
for this project can be accessed at \url{https://codeberg.org/jinensetpal/context-regularized-cross-entropy/}.

\paragraph{Asset Licenses.}
We used the following existing assets, listed with their respective licenses:
\begin{itemize}[leftmargin=*]
    \item The Waterbirds dataset \citep{sagawa2019distributionally} is licensed under MIT.
    \item The Spawrious dataset \citep{lynch2023spawrious} is licensed under CC0-1.0.
    \item The ImageNet dataset \citep{deng2009imagenet} is licensed only for non-commercial purposes.
    \item The ImageNet-BG dataset \citep{Piekarek_2025_ICCV} is published under a public domain license. 
    \item The Hard-ImageNet dataset \citep{moayeri2022hard} is published under CC-0.
    \item Model weights for the Masked AutoEncoder (MAE) \citep{he2022masked} trained on the Large-Scale Unsupervised Segmentation (LUSS) \citep{gao2021luss} dataset is licensed under CC BY-NC 4.0.
    \item Model weights for ABNs \citep{fukui2019attention} is licensed under MIT.
    \item SAM3 \citep{carion2026sam} is published under a custom SAM License which permits the academic use of their tool.
    \item The \texttt{torchvision} library is published under a BSD 3-Clause License.
\end{itemize}

\clearpage


\end{document}